\documentclass[final,5p,times,twocolumn]{elsarticle}

\usepackage{amsmath,amssymb,amsfonts}
\usepackage{booktabs}
\usepackage{multirow}
\usepackage{xcolor}
\usepackage{setspace}
\usepackage{enumitem}
\usepackage[ruled,vlined,linesnumbered]{algorithm2e}
\usepackage{pifont}
\usepackage{tabularx}
\usepackage{array}
\usepackage{graphicx}
\usepackage{setspace}
\usepackage{url}
\usepackage{hyperref}
\usepackage[table]{xcolor}
\usepackage{makecell}

\newcommand{\inscat}{\textsc{InsCAT}}
\newcommand{\sica}{\textsc{SICA}}
\newcommand{\ropo}{\textsc{ROPO}}
\newcommand{\pbcat}{\textsc{PBCAT}}

\newcommand{\calL}{\mathcal{L}}
\newcommand{\bR}{\mathbb{R}}
\newcommand{\zadv}{z_\text{adv}}
\newcommand{\zcle}{z_\text{cle}}
\newcommand{\zneg}{z_\text{neg}}
\newcommand{\sg}{\mathrm{sg}}
\definecolor{lightgrayrow}{gray}{0.85}

\begin{document}

\begin{frontmatter}

\title{Detectors Learn the Wrong Thing: Shortcut-Resistant Adversarial Training Against Physically Realizable Attacks}

\author[label1,label2]{Yuanhao Huang}
\ead{yuanhao\_huang@buaa.edu.cn}

\author[label1,label3]{Yilong Ren\corref{cor1}}
\ead{yilongren@buaa.edu.cn}

\author[label4]{Jinlei Wang}
\ead{20231800822@imut.edu.cn}

\author[label1,label2]{Xuesong Bai}
\ead{xs\_bai@buaa.edu.cn}

\author[label5]{Zheng Zhang}
\ead{xs\_bai@buaa.edu.cn}

\author[label1,label3]{Haiyang Yu\corref{cor1}}
\ead{hyyu@buaa.edu.cn}


\cortext[cor1]{Corresponding author}

\cortext[cor2]{Funding: This work was supported by the Fundamental and Interdisciplinary Disciplines Breakthrough Plan of the Ministry of Education of China(No.JYB2025XDXM123), Shandong Provincial Key R\&D Program, [Grant No. 2025CXGC010109], State Key Lab of Intelligent Transportation System, and fundamental research funds for the central universities.}

\affiliation[label1]{organization={School of Transportation Science and Engineering, Beihang University},
postcode={100191}, state={Beijing}, country={P.R China}}

\affiliation[label2]{organization={State Key Lab of Intelligent Transportation System},
postcode={100191}, state={Beijing}, country={P.R China}}

\affiliation[label3]{organization={Zhongguancun Laboratory},
postcode={100191}, state={Beijing}, country={P.R China}}

\affiliation[label4]{organization={School of Systems Science and Engineering, Sun Yat-Sen University},
postcode={510006}, state={Guangzhou}, country={P.R China}}

\affiliation[label5]{organization={Shandong Sino-Aisa Tire Proving Ground Co.,Ltd.},
postcode={200021}, state={Shandong}, country={P.R China}}

\begin{abstract}
AI-enabled visual perception systems are increasingly deployed in intelligent transportation infrastructure and autonomous vehicle-related applications. However, physically realizable adversarial appearances pose a significant reliability challenge for these safety-critical systems. Adversarial training is effective, but repeated co-occurrence between adversarial texture and positive person instances can cause detectors to treat the texture itself as evidence of object presence, forming a patch texture shortcut. The detector may then treat texture as evidence for the target, causing false detections on texture only inputs and weakening cross attack generalisation. We propose InsCAT, an instance-level contrastive adversarial training framework that prevents detectors from using adversarial texture as an independent decision cue. SICA aligns adversarial person features with matched clean features and separates them from texture only negatives, while ROPO and Guard maintain online attack pressure and coordinate training. We evaluate eight independently generated attack textures on rendered nuScenes, INRIAPerson, printed garments, and three detector families. InsCAT achieves an average attack AP of $82.3\%$ on rendered nuScenes, exceeding the strongest baseline by $11.1$ points. Relative to AT-Mix, texture FPR decreases from $46.9\%$ to $7.3\%$. Physical tests yield an F1 score of $96.6\%$ and an FPR of $1.8\%$. Consistent gains across separately trained detectors demonstrate applicability across architectures with direct inference. The findings show that robust physical detection depends on preserving target related evidence while preventing adversarial texture from becoming an independent decision cue. The code is avaliable at \url{https://github.com/Huangyh98/InsCAT.git}.
\end{abstract}


\end{frontmatter}


\section{Introduction}
\label{sec:intro}

Object detection is a fundamental perception capability in intelligent systems, enabling machines to localise and recognise objects in the physical world. Pedestrian perception is a fundamental capability in intelligent transportation infrastructure, surveillance systems, and autonomous vehicle deployment, where unreliable detection may compromise downstream safety decisions. Modern detectors built on convolutional and Transformer architectures achieve strong accuracy but remain vulnerable to adversarial examples designed to induce confident mispredictions~\cite{huang2024advswap}. Unlike digital attacks confined to pixel space, physically realisable attacks can be printed and deployed in the real world~\cite{xu2020adversarial}. For person detection, the attacks range from localised patches attached to clothing~\cite{thys2019fooling} to optimised full body textures worn as garments~\citep{hu2022adversarial,hu2023physically}. Such attacks can suppress person detections across changes in viewpoint, camera, and illumination, making robust defence an important requirement.

Defences against physically realisable attacks generally follow two routes. Input purification locates and neutralises adversarial regions before detection, but becomes less reliable when the perturbation is distributed across full body clothing textures~\cite{kumar2025unified}. Adversarial training instead incorporates attacked samples into detector optimisation and directly shapes the learned representation~\cite{madry2017towards}. Most adversarial training research focuses on small norm perturbations in image classification~\cite{zhang2019towards}. Physical adversarial training differs because visible attack patterns are embedded in positive target instances and remain associated with their category labels throughout optimisation. The central question is therefore not only whether attacked samples are included during training, but also what evidence the detector learns from them.

Existing adversarial training methods against physically realisable attacks often show limited cross attack generalisation, as robustness learned from one attack texture does not reliably transfer to independently generated textures. PBCAT improves attack coverage by combining local patches with full image perturbations and broadening the training distribution at the data level~\cite{li2025pbcat}. This strategy expands the range of observed attacks but does not explicitly constrain which visual cues are encoded by the detector.  
Label leaking shows that adversarial training can exploit regularities introduced during adversarial example generation~\cite{kurakin2016adversarial}. Shortcut learning and simplicity bias further show that models favour predictive cues that provide an easier route to the training objective~\cite{geirhos2020shortcut,shah2020pitfalls}. The findings motivate examining whether an optimised physical pattern can itself become evidence for the target category. In this case, the detector does not merely learn the presence of a person under adversarial appearance changes; it may learn the adversarial texture as a predictive cue for person detection. This texture-driven evidence represents the wrong thing learned during adversarial training.

\begin{figure}[t!]
  \centering
  \includegraphics[width=8.6cm]{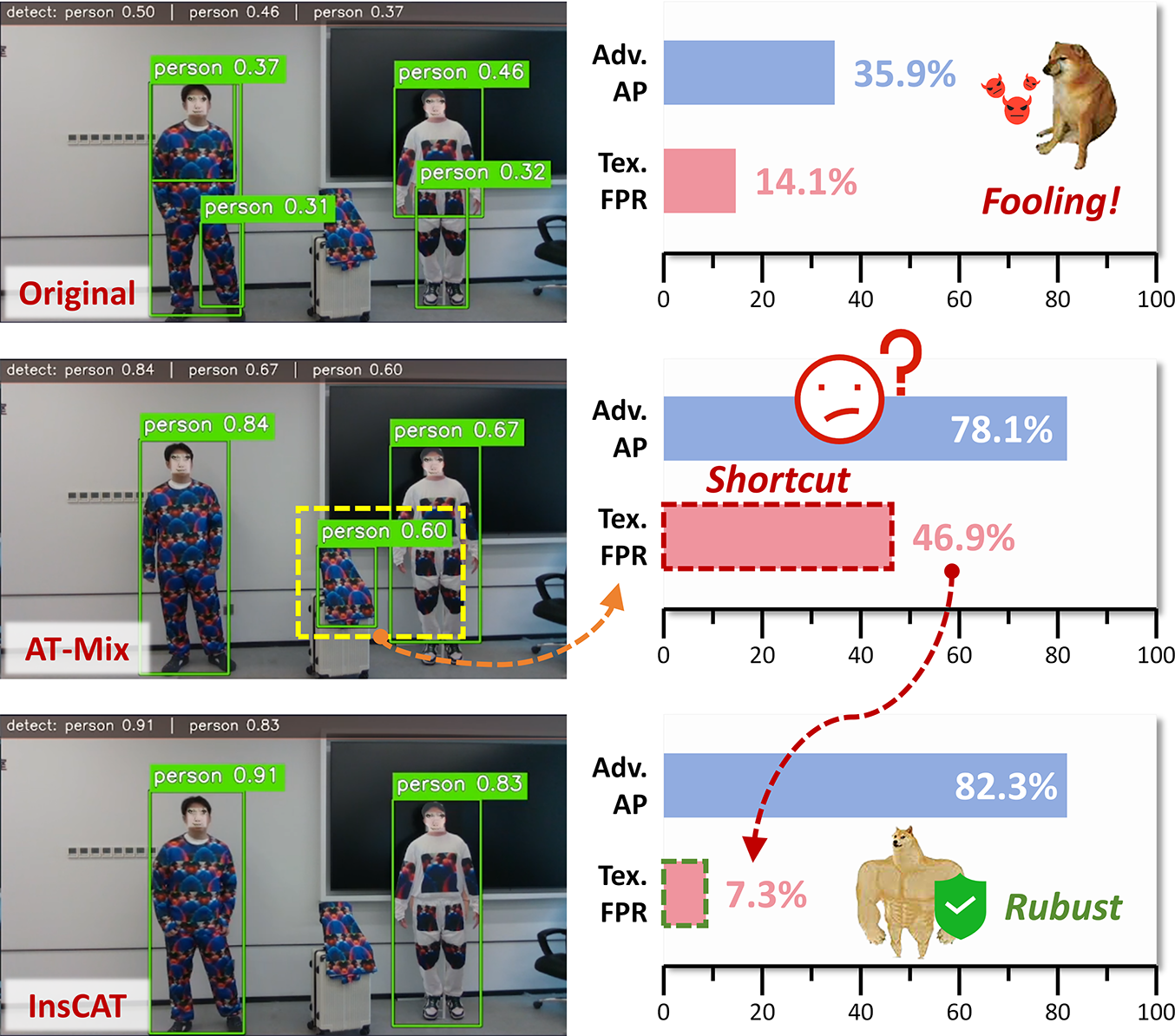}
  \caption{\textbf{The patch texture shortcut.} The undefended detector, AT-Mix, and InsCAT are evaluated on adversarially clothed persons using attack AP and on human-free texture inputs using texture FPR.}
  \label{fig:shortcut}
  \vspace{-4 mm}
\end{figure}

Physical adversarial training repeatedly presents targets carrying optimised attack patterns as positive samples~\citep{li2025pbcat,bayer2026higher,strack2024defending}. Because the patterns are explicitly optimised against the detector, they provide strong and easily learned responses within the target region~\citep{ilyas2019adversarial,geirhos2020shortcut,shah2020pitfalls}. When such patterns repeatedly appear with positive person instances during adversarial training, the detector can exploit the texture itself as an easier decision cue than person-related semantic evidence. The detector can therefore reduce the training loss by treating the attack pattern itself as evidence for object presence. In the person detection setting studied here, our feature diagnostics show that adversarial person representations become more similar to texture only representations, while texture only inputs trigger confident person predictions without a human target. We call this failure the patch texture shortcut. It allows strong detection on adversarially textured persons to coexist with texture triggered false positives and weaker transfer across texture appearances. Fig.~\ref{fig:shortcut} illustrates this contradiction. AT-Mix substantially improves adversarial AP but produces a high texture false positive rate on human free inputs, whereas \inscat{} maintains strong adversarial detection while suppressing texture triggered predictions.

We propose InsCAT, an Instance Level Contrastive Adversarial Training framework that suppresses this texture-driven shortcut. Its core module, \sica{} (Structure-Invariant Contrastive Alignment), encourages the representation of an adversarial person to remain consistent with its matched clean counterpart while separating it from a texture-only negative generated from the same online texture. It aligns the adversarial person representation with the clean person representation while separating it from texture-related evidence. To support this feature regulation under online physical adversarial training, \ropo{} (Rendering Amortised Online Patch Optimisation) maintains an evolving global clothing texture through joint 2D and 3D optimisation and amortises differentiable rendering through sample reuse. An adversarial guard stabilises the training trajectory by coordinating the clean detection, adversarial detection, and contrastive objectives according to validation performance. Together, these components maintain an evolving adversarial signal while reducing the detector's reliance on texture-only cues.

The main contributions are summarised as follows.

\begin{enumerate}[nosep,leftmargin=*]
\item We identify a hidden reliability issue in AI-based perception systems, where adversarial training can unintentionally encourage detectors to rely on attack-specific visual patterns. In person detection, this shortcut allows clothing texture to become independent evidence for person presence. We introduce the texture false positive rate on human-free texture inputs and instance-level feature relationships to diagnose this failure.
\item We propose \sica{}, an instance-level feature regulation mechanism that aligns adversarial person representations with their matched clean counterparts and separates them from texture-only negatives. Building on this mechanism, \inscat{} integrates \ropo{} for rendering-amortised online attack generation and an adversarial guard for training coordination, enabling shortcut-resistant physical adversarial training.
\item We evaluate \inscat{} against eight independently generated attack textures on rendered nuScenes and INRIAPerson, and further validate it using printed garments and three detector architectures. The results show consistent adversarial robustness gains together with lower texture false positive rates and stronger separation from texture-only representations.
\end{enumerate}

The remainder of this paper is organised as follows. Section~\ref{sec:related} reviews related work, Section~\ref{sec:method} introduces \inscat{}, Section~\ref{sec:experiments} presents the experiments, Section~\ref{sec:discussion} discusses the main findings and limitations, and Section~\ref{sec:conclusion} concludes the paper.

\section{Related Work}
\label{sec:related}

\subsection{Physical Adversarial Attacks on Object Detection}

Physical adversarial attacks on object detectors have developed along three main dimensions: spatial coverage, visual naturalness, and physical transferability. AdvPatch and AdvT-shirt established bounded printed patches and deformable wearable attacks for suppressing person detection~\citep{tom2017adversarial,xu2020adversarial}. AdvTexture and AdvCaT expanded the attack to full body clothing surfaces, improving robustness to viewpoint changes, garment deformation, and natural appearance~\citep{hu2022adversarial,hu2023physically}. NatPatch introduces generative priors for natural looking patterns, while T-SEA and TRDPatch strengthen transfer across detector architectures and dynamic physical conditions~\citep{hu2021naturalistic,huang2023t,wang2025transferable}. AdvReal combines 2D and 3D optimisation to improve cross viewpoint robustness~\citep{huang2026advreal}, while AdvSerial extends physical attacks to continuous frames~\citep{huang2026advserial}. Together, these methods produce attack families that differ substantially in spatial extent, appearance, deformation, model transfer, viewpoint, and temporal behaviour. The diversity motivates evaluating defence robustness across independently generated attacks rather than against a single training texture.

\subsection{Defences against Physically Realisable Attacks}

One line of defence removes the adversarial pattern before detection. LGS and Jedi identify suspicious regions from local image statistics~\cite{naseer2019local,tarchoun2023jedi}, while PAD exploits semantic independence and spatial heterogeneity~\cite{jing2024pad}. SAC, PatchZero, and NAPGuard learn to segment or detect adversarial regions~\cite{liu2022segment,xu2023patchzero,wu2024napguard}. The methods differ in how the attack region is identified, but all rely on separating it from the target. This becomes difficult for full body textures, where the optimised pattern is distributed across the target surface.

Adversarial training instead learns directly from attacked samples~\citep{madry2017towards}. Strong iterative attacks improve robustness but introduce substantial training cost, motivating FreeAT to reuse gradient computation across attack and model updates~\citep{shafahi2019adversarial}. Physical adversarial training extends this paradigm from norm bounded perturbations to visible and spatially constrained attacks. Existing methods optimise rectangular occlusions, patch locations, or universal patches during training~\citep{wu2019defending,rao2020adversarial,metzen2021meta}. PBCAT broadens the training threat model through composite local and global perturbations for object detection~\citep{li2025pbcat}. The methods advance the construction, efficiency, diversity, and transformation robustness of the training attack. Our work shifts the focus to the instance level evidence learned from adversarially textured positive samples.

\subsection{Shortcut Learning and Spurious Correlations}

Shortcut learning describes the tendency of neural networks to exploit predictive regularities that satisfy the training objective without capturing the intended semantics~\cite{geirhos2020shortcut}. Such behaviour appears in both standard and adversarial learning. Background correlations can dominate visual recognition decisions~\citep{xiao2020noise}, while non robust features remain highly predictive despite being brittle under adversarial perturbations~\citep{ilyas2019adversarial}. Label leaking further shows that adversarial training can absorb regularities introduced by the attack generation process~\citep{kurakin2016adversarial}. Together, these studies show that data composition and optimisation jointly determine which visual cues become part of the decision rule.

Feature reliance studies provide a direct basis for analysing these cues. Geirhos et al.~\citep{geirhos2018imagenet} show that ImageNet trained CNNs often follow texture in shape texture conflict images. Burgert et al.~\citep{burgert2026imagenet} revisit this conclusion through controlled feature suppression and find that standard CNNs rely strongly on local shape cues. These results highlight the importance of isolating individual cues rather than inferring feature dependence from task performance alone. Our work examines a training induced dependence in physical adversarial defence, where optimised texture is repeatedly coupled with the target label. We identify the resulting patch texture shortcut through texture only inputs and instance level feature relationships.

\subsection{Contrastive Learning in Adversarial Robustness}

Contrastive learning builds invariant representations by aligning different views of the same input while separating representations from different samples. SimCLR and MoCo establish this paradigm for self supervised visual representation learning~\cite{chen2020simple,he2020momentum}. ACL and RoCL incorporate adversarial views into contrastive pretraining to improve representation robustness under norm bounded perturbations~\citep{jiang2020robust,kim2020adversarial}. AdvCL further studies how adversarial robustness can be preserved from contrastive pretraining to downstream fine tuning~\cite{fan2021does}. These methods constrain global image representations and are primarily developed for image classification.

Instance level representations provide a more targeted basis for analysing physical adversarial effects in object detection. Whole image features combine the attacked target with background content and other objects, making it difficult to isolate the relationship between the target and the adversarial texture. FSCE shows that proposal level contrastive learning can capture instance specific representations within detection models~\citep{sun2021fsce}. Following this idea, \inscat{} applies contrastive regulation to ground truth aligned RoI features, aligning the adversarial target with its matched clean counterpart and separating it from the corresponding texture only representation.

\section{Methodology}
\label{sec:method}

\subsection{Threat Model and Defense Objective}
\label{sec:formulation}

This section defines the physically realizable clothing-texture threat considered in this work and introduces the rendered-triplet objective underlying the outer detector optimisation. The complete training objective, including clean detection, buffered adversarial detection, and the rendering-amortised update schedule, is presented in Sec.~\ref{sec:method:schedule}.

\textbf{Threat model.}
We consider physically realizable evasion attacks that suppress the person class of an object detector. The adversary designs a printable clothing texture $\mathbf{p}\in\mathcal{P}$, where $\mathcal{P}=[0,1]^{H_t\times W_t\times 3}$ denotes the feasible UV texture space. The texture is mapped onto the garment surface of a 3D human mesh $\mathcal{M}$ and projected into a scene image through a physically parameterised renderer $\mathcal{R}$.
\begin{equation}
\label{eq:render}
I_{\mathrm{adv}}=\mathcal{R}(\mathcal{M},\mathbf{p},\mathbf{c})
\end{equation}
where $\mathbf{c}=(\phi,\mathbf{l},B,\mathbf{s})$ specifies the viewpoint azimuth $\phi$, lighting condition $\mathbf{l}$, background $B$, and body shape and pose $\mathbf{s}$. For each matched clean and adversarial rendering, only the garment appearance is modified while the human geometry and scene condition remain unchanged.

Let $s_{\theta}(I)=\max_{q\in\mathcal{Q}_{\theta}(I)}P_{\theta}(\mathrm{person}\mid q)$ denote the maximum person detection score over the candidate predictions $\mathcal{Q}_{\theta}(I)$ produced by detector $f_{\theta}$. The patch generation objective suppresses the person response while regularising the spatial variation of the printable texture.
\begin{equation}
\label{eq:patch-obj}
\mathcal{J}_{\mathrm{patch}}(\mathbf{p};\theta)=\mathbb{E}_{\mathbf{c}}\!\left[s_{\theta}\!\left(\mathcal{R}(\mathcal{M},\mathbf{p},\mathbf{c})\right)\right]+\lambda_{\mathrm{tv}}\calL_{\mathrm{TV}}(\mathbf{p})
\end{equation}
where $\lambda_{\mathrm{tv}}>0$ controls the strength of the total variation regularisation.

The adversarial texture is obtained by minimising the person response and the texture variation penalty.
\begin{equation}
\label{eq:attack-obj}
\mathbf{p}^{*}=\arg\min_{\mathbf{p}\in\mathcal{P}}\mathcal{J}_{\mathrm{patch}}(\mathbf{p};\theta)
\end{equation}

The attack utility is defined as
\begin{equation}
\label{eq:attack-utility}
\calL_{\mathrm{attack}}(\mathbf{p};\theta)=-\mathcal{J}_{\mathrm{patch}}(\mathbf{p};\theta)
\end{equation}

The inner attack optimisation is then written as
\begin{equation}
\label{eq:inner}
\mathbf{p}^{*}=\arg\max_{\mathbf{p}\in\mathcal{P}}\calL_{\mathrm{attack}}(\mathbf{p};\theta)
\end{equation}

During training, the inner adversary updates a shared clothing texture using gradients from the current detector. The evaluation textures are generated independently and are not used during detector training.

\textbf{Defense objective.}
Given the adversarial texture produced by the inner optimisation, the detector is trained to preserve person detection under adversarial appearance changes and to reduce its dependence on texture related cues.
\begin{equation}
\label{eq:minmax}
\min_{\theta}\mathbb{E}_{(\mathbf{c},y)\sim\mathcal{D}_{\mathrm{rend}}}\!\left[\calL_{\mathrm{det}}\!\left(\theta;I_{\mathrm{adv}},y\right)+\alpha_{\mathrm{s}}\calL_{\sica}\!\left(\theta;I_{\mathrm{adv}},I_{\mathrm{cle}},I_{\mathrm{neg}}\right)\right]
\end{equation}
where
\begin{equation}
\label{eq:sample-triplet}
I_{\mathrm{adv}}=\mathcal{R}(\mathcal{M},\mathbf{p}^{*},\mathbf{c}),\qquad I_{\mathrm{cle}}=\mathcal{R}(\mathcal{M},\mathbf{p}_{\mathrm{cle}},\mathbf{c}),\qquad I_{\mathrm{neg}}=x_{\mathrm{neg}}(\mathbf{p}^{*})
\end{equation}

Here, $p_{\mathrm{cle}}$ denotes the clean garment texture, while $x_{\mathrm{neg}}(p^*)$ denotes a texture-only negative generated by compositing the current adversarial texture onto a blank background canvas without any human pixels. The clean and adversarial renderings share the same scene condition and provide matched person instances for feature alignment.

\subsection{Framework Overview}

\begin{figure*}[t]
\centering
\includegraphics[width=16cm]{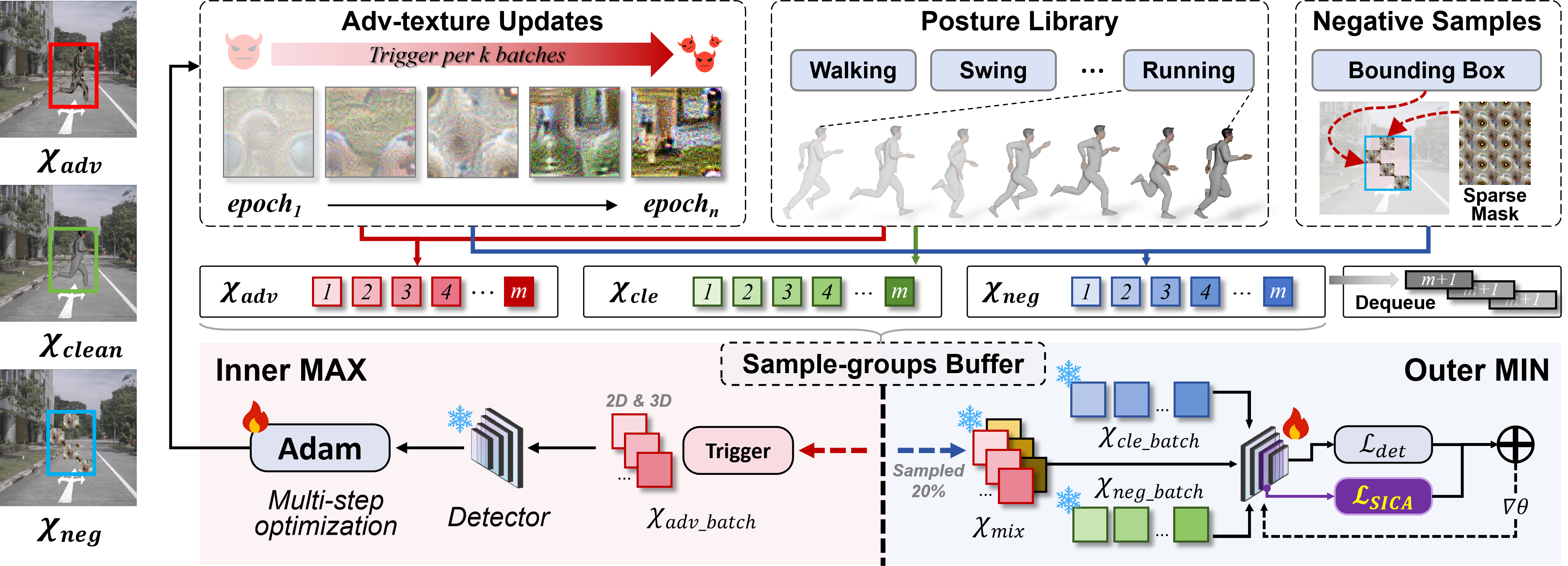}
\caption{\textbf{The \inscat{} framework.} The top row contains online adversarial texture updates, a posture library covering walking, running, crossing, and sneaking, and dynamically generated human free texture negatives. The middle row shows how the adversarial, clean, and negative sample groups $\chi_{\mathrm{adv}}/\chi_{\mathrm{cle}}/\chi_{\mathrm{neg}}$ are maintained in the sample groups buffer. In the bottom row, \emph{Inner MAX} freezes the detector and updates the adversarial texture through the 2D and 3D attack branches. \emph{Outer MIN} updates the detector using buffered adversarial samples, the detection loss $\calL_{\mathrm{det}}$, and the contrastive loss $\calL_{\sica}$.}
\label{fig:overview}
\vspace{-4 mm} 
\end{figure*}

\inscat{} embeds an online min-max patch optimiser into a standard object-detector training loop, and Fig.~\ref{fig:overview} gives the overview. As shown in the top row, a single global adversarial texture is updated online and rendered onto a posture library (walking, running, crossing, sneaking), while a human-free negative is synthesised from the same texture so that the patch appears with no human body. The middle row collects the resulting renderings into adversarial, clean, and negative sample groups $\chi_\text{adv}/\chi_\text{cle}/\chi_\text{neg}$, held in a render buffer that decouples the expensive rendering from the fast detection updates. The bottom row is the min-max alternation that drives optimisation. \emph{Inner MAX} freezes the detector and strengthens the patch with Adam over a 3D rendering branch and a 2D image branch, whereas \emph{Outer MIN} unfreezes the detector, mixes buffered adversarial samples into the batch, and jointly back-propagates the detection loss $\calL_\text{det}$ and the \sica{} contrastive loss $\calL_\sica$ to update $\theta$.

To amortise the dominant rendering cost, the attack side is refreshed only once every $f_i$ detection batches. Between refreshes the detector keeps training on buffered renderings, and outer batches mix buffered adversarial samples at a ratio $\rho_\text{adv}$. A clean-floor-gated \emph{robust phase} and an \emph{adversarial guard} adjust the loss weights once clean accuracy is safe, so that the optimiser does not drift back to pure clean detection in late training. The following subsections proceed along this data flow, covering the rendering-amortised online patch optimisation that produces the adversarial training signal (Section~\ref{sec:ropo}), the dynamic generation of texture-only negatives (Section~\ref{sec:neg}), the instance-level contrastive alignment that consumes the resulting triplets (Section~\ref{sec:method:sica}), and the overall objective and training schedule (Section~\ref{sec:method:schedule}).

\subsection{Rendering Amortised Online Patch Optimization}
\label{sec:ropo}

\ropo{} maintains a shared adversarial texture throughout training and reduces repeated differentiable rendering by reusing recent rendered samples across detector updates. It follows the amortisation principle of FreeAT~\cite{shafahi2019adversarial} and applies it to the rendering process. FreeAT replays training batches and updates adversarial perturbations and model parameters using gradients obtained from the same backward pass. \ropo{} periodically refreshes the adversarial texture and rendered samples while the detector continues training with samples stored in the render buffer. The method contains three components.

\begin{itemize}[nosep,leftmargin=*]

\item \textbf{Render buffer.} A FIFO queue with capacity $m$ stores the latest adversarial renderings and their detection labels. The adversarial texture and rendered samples are refreshed every $f_i$ detector updates. Between two refreshes, detector training reuses samples from the buffer and distributes the rendering cost across multiple training batches.

\item \textbf{Single global online texture.} \ropo{} maintains one globally shared UV texture throughout training. The same texture is optimised across different viewpoints, poses, lighting conditions, and backgrounds. This setting follows the physical constraint that one printable garment texture must remain effective under varying observation conditions.

\item \textbf{Scene consistent rendering.} The scene condition $\mathbf{c}$ is sampled and cached before each texture update. The clean and adversarial instances are rendered using the same viewpoint, lighting, background, body shape, and pose. The resulting image pair differs only in garment appearance and provides matched person instances for \sica{}.

\end{itemize}

\textbf{Inner maximisation.}
At each ROPO refresh, the detector parameters are frozen and the shared texture is updated for $n_p$ Adam steps. The practical attack utility extends Eq.~\eqref{eq:attack-utility} with an auxiliary 2D attack branch.
\begin{equation}
\label{eq:attack}
\calL_{\mathrm{attack}}^{\mathrm{ROPO}}(\mathbf{p};\theta)=-s_{\theta}(I_{\mathrm{3D}})-s_{\theta}(I_{\mathrm{2D}})-\lambda_{\mathrm{tv}}\calL_{\mathrm{TV}}(\mathbf{p})
\end{equation}
where $I_{\mathrm{3D}}=\mathcal{R}(\mathcal{M},\mathbf{p},\mathbf{c})$ denotes the physically rendered image, $I_{\mathrm{2D}}$ denotes the image generated by differentiable texture compositing, and $\calL_{\mathrm{TV}}(\mathbf{p})=TV_h(\mathbf{p})+TV_w(\mathbf{p})$. The two detection terms are assigned equal weights in the current implementation.

Maximising Eq.~\eqref{eq:attack} suppresses the person detection scores in both branches and reduces excessive spatial variation in the adversarial texture. The optimisation is implemented by minimising $-\calL_{\mathrm{attack}}^{\mathrm{ROPO}}$ with Adam.

The 3D branch implements the physical threat defined in Section~\ref{sec:formulation}. It renders the textured human model under the sampled scene condition and propagates the detector gradient through the differentiable renderer. The 2D branch provides an auxiliary attack signal using person images sampled exclusively from the 614 positive images in the INRIAPerson training split, denoted as Train/pos. The Test/pos split is not accessed during detector training or online texture optimisation. Each image is resized to the detector input resolution, and a differentiably scaled copy of the shared texture is composited into the person region through a non-inplace operation. The auxiliary branch increases image domain diversity and provides additional gradients for online texture optimisation.

\textbf{Physical rendering pipeline.}
The renderer $\mathcal{R}$ in Eq.~\eqref{eq:render} is implemented with PyTorch3D using multiple human motion sequences. The motion sequences are sampled in a round robin order to maintain a balanced pose distribution. Each rendering step samples an azimuth $\phi\sim\mathcal{U}(-180^\circ,180^\circ)$, a point light, and a background crop from nuScenes~\cite{caesar2020nuscenes}.

The adversarial texture is applied through tiled UV mapping with tile scale $s_t$ as specified in Section~\ref{sec:experiments:setup}. Texture blocks are densely repeated across the shirt and trouser UV atlases. A close range rendering mode generates partially visible persons containing the upper body or part of the lower body. These samples reflect the truncation patterns observed during close range physical capture. The same rendering configuration is used to generate the training samples and the nuScenes evaluation scenes while the evaluation attack textures remain independent from the texture optimised during training.

\subsection{Dynamic Negative Sample Generation}
\label{sec:neg}

The negative contains no human structure and consists only of the adversarial texture placed on a blank background. Given the updated patch $p^\star$ after inner maximisation, the negative is constructed in five steps. The patch is first scaled to a random fraction of the reference bounding-box width while preserving its aspect ratio, and is then tiled into a mosaic covering the reference box. The mosaic is cropped to the box size at a random offset, after which a sparse random mask retaining a fraction of its cells is applied. The resulting sparse texture is finally composited onto a blank background canvas. The bounding box is used only to determine the spatial extent of the texture region and does not contain any human pixels.

The negative is updated together with the current patch so that its adversariality stays matched to training progress. To avoid collapsing to a single-texture negative, with probability $p_s$ the negative is generated from a snapshot drawn from the same online patch trajectory rather than from the current patch. This trajectory-snapshot negative requires neither a held-out patch cache nor external attack textures. It spans only the texture states that the online patch has already visited, discouraging dependence on the texture at any single instant.

\begin{figure*}[t]
  \centering
  \includegraphics[width=16cm]{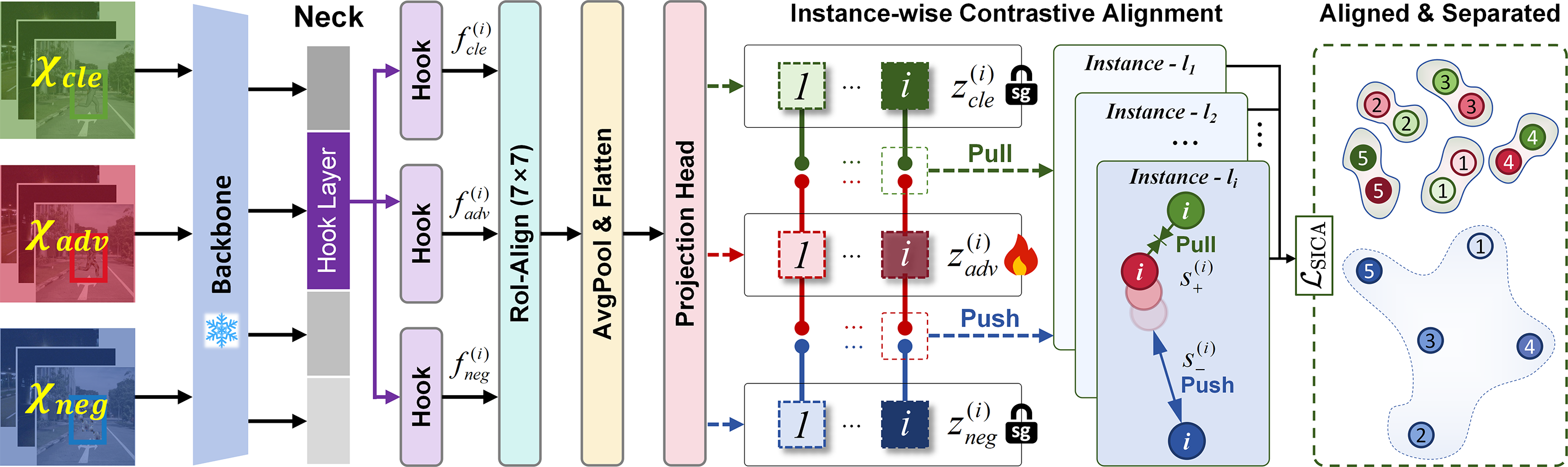}
  \caption{\textbf{\sica{} instance-level contrastive alignment.} The three inputs $\chi_\text{cle}/\chi_\text{adv}/\chi_\text{neg}$ share a frozen backbone. A forward hook extracts the selected mid-level feature map, and RoIAlign followed by average pooling and a projection head produces the normalised embeddings $\zcle^{(i)}/\zadv^{(i)}/\zneg^{(i)}$. Each person instance forms an independent triplet. Within each SICA update, stop-gradient treats the clean and texture-negative embeddings as fixed targets, so gradients from $\calL_\sica$ propagate through the adversarial branch. The objective aligns $\zadv$ with $\zcle$ and separates it from $\zneg$, encouraging consistent person representations with reduced dependence on adversarial texture cues.}
  \label{fig:sica}
  \vspace{-4 mm} 
\end{figure*}

\subsection{Structure-Invariant Contrastive Alignment (SICA)}
\label{sec:method:sica}

With the adversarial-clean pairs of Section~\ref{sec:ropo} and the texture-only negatives of Section~\ref{sec:neg} in hand, \sica{} imposes the instance-level feature constraint that forms the core mechanism of \inscat{}. Here, structure-invariant refers to consistency of the matched person representation when only the garment appearance changes. The clean and adversarial renderings share the same human geometry and scene conditions, allowing their RoI features to form a positive pair that retains person-related information. The texture-only representation provides the corresponding negative and discourages reliance on the attack pattern. \sica{} constructs this constraint independently for each person instance, as illustrated in Fig.~\ref{fig:sica}.

\textbf{Alignment layer.}
\label{sec:method:layer}
Detection heads encode instances through architecture specific mechanisms. YOLOv5 assigns predictions to anchors, Faster R-CNN operates on region proposals, and DN-DETR represents objects through matched queries. Direct alignment at the prediction head would therefore bind SICA to detector specific correspondence and output structures. We attach SICA to a late spatial feature map before these head specific transformations. Ground truth boxes then define consistent RoI features for the same person instance across clean, adversarial, and texture only inputs.

Late backbone and neck features retain spatial correspondence while carrying stronger object semantics than shallow features. They also expose adversarial feature shifts before classification, regression, and query matching convert them into final predictions~\citep{metzen2017detecting}. Feature alignment at this stage has improved clean AP and adversarial robustness in both one stage and two stage object detectors~\citep{xu2021using}. SICA therefore operates on a deep backbone or neck feature immediately before the task specific detection head. For YOLOv5n, the layer study in Tab.~\ref{tab:layer_select} selects P5 as the default attachment layer.

\textbf{RoI feature extraction.}
A forward hook on the selected mid-level layer captures the feature map $\mathbf{F}\in\bR^{C\times H_F\times W_F}$ from each forward pass. For every ground-truth box, we extract a fixed-size instance feature using RoIAlign with spatial scale $H_F/H_{\mathrm{in}}$, followed by adaptive average pooling. The pooled feature is passed through a projection head composed of a two-layer MLP with BatchNorm and ReLU, and the output is $\ell_2$ normalised onto a unit hypersphere. BatchNorm stabilises the projected feature distribution during optimisation. The backbone is frozen, while the neck, detection head, and projection head are updated jointly. Each ground-truth instance defines a triplet $(\zadv,\zcle,\zneg)$ that aligns the adversarial representation with its matched clean counterpart and separates it from the corresponding texture-only representation.

\textbf{Contrastive objective.}
Let $s_\text{pos}^{(i)}=\zadv^{(i)\top}\zcle^{(i)\,\sg}$ and $s_\text{neg}^{(i)}=\zadv^{(i)\top}\zneg^{(i)\,\sg}$ be the positive and negative cosine similarities, where $\sg$ denotes stop-gradient. The legacy form is a temperature-scaled InfoNCE loss
\begin{equation}
\label{eq:infonce}
\begin{aligned}
  \calL_\sica^\text{NCE} ={}& \frac{\lambda}{N}\sum_{i=1}^{N}
  \Big[-\frac{s_\text{pos}^{(i)}}{\tau} + \log\!\big(e^{s_\text{pos}^{(i)}/\tau}
      + e^{s_\text{neg}^{(i)}/\tau}\big)\Big],
\end{aligned}
\end{equation}
with temperature $\tau$ and weight $\lambda$. The InfoNCE form can let $s_\text{pos}$ and $s_\text{neg}$ rise together, so the adversarial feature moves closer to both the clean anchor and the texture negative without forming a sufficient relative margin. We therefore adopt a gap-aware form as the main method, which optimises the relative margin directly through an explicit hinge:
\begin{equation}
\label{eq:gap}
  \calL_\text{gap} = \frac{1}{N}\sum_i
    \max\!\left(0,\;\delta + s_\text{neg}^{(i)} - s_\text{pos}^{(i)}\right),
\end{equation}
\begin{equation}
\label{eq:sica-gap}
\begin{aligned}
  \calL_\sica ={}& \lambda\Big[
    w_\text{pos}\tfrac{1}{N}\!\sum_i\!\left(1 - s_\text{pos}^{(i)}\right)
    + w_\text{gap}\calL_\text{gap} \\[2pt]
    &+ w_\text{neg}\tfrac{1}{N}\!\sum_i\!\max\!\left(0, s_\text{neg}^{(i)} - m_\text{neg}\right)\Big],
\end{aligned}
\end{equation}
where $\lambda$ is the fixed base scale of the complete SICA objective. The weights $w_\text{pos}$, $w_\text{gap}$, and $w_\text{neg}$ determine the relative contributions of the positive alignment, gap, and negative separation terms. The coefficient $\alpha_s$ is adjusted by the training scheduler and controls the phase-dependent contribution of SICA to detector optimisation. The resulting SICA contribution is scaled by $\alpha_s\lambda$. The margin $\delta$ defines the required separation between the positive and negative similarities, while $m_\text{neg}$ limits the permitted similarity to the texture-negative embedding.

\textbf{Stop-gradient.}
Within each SICA update, the clean and texture-negative embeddings are detached from $\calL_\sica$ and treated as fixed alignment targets. This prevents the contrastive objective from moving both embeddings in each pair during the same backward pass. Gradients from $\calL_\sica$ therefore propagate only through the adversarial branch, pulling the adversarial embedding toward the clean target and pushing it away from the texture-negative target. These targets remain fixed within the current contrastive update, while the shared detector parameters continue to be updated by the complete training objective.

\subsection{Overall Objective and Training Schedule}
\label{sec:method:schedule}

\ropo{} implements the bilevel adversarial training objective defined by Eqs.~\eqref{eq:inner} and~\eqref{eq:minmax} through alternating optimisation. The inner step updates $\mathbf{p}$ under the attack utility in Eq.~\eqref{eq:attack} while keeping $\theta$ fixed. The outer step fixes $\mathbf{p}$ and updates $\theta$ using the complete training objective defined below. The global training loss consists of the clean detection loss, the buffered adversarial detection loss, and the rendered adversarial and SICA losses computed at each \ropo{} refresh.
\begin{equation}
\label{eq:total}
\begin{aligned}
\calL_\text{total}={}&w_c\calL_\text{cle}
+w_a\calL_\text{adv-buf}\\
&+\mathbf{1}_\ropo\!\left(
\calL_\text{adv-rend}
+\alpha_s\calL_\sica
\right).
\end{aligned}
\end{equation}

A clean floor $\phi_c$ is defined as the epoch zero clean validation metric minus a tolerance $\epsilon_c$. The adversarial validation metric $m_a$ is evaluated using validation renderings generated with the current globally shared texture $\mathbf{p}$ produced by the inner optimisation. The independently generated attack textures used for final evaluation are not involved in the training schedule. The weights $w_c$, $w_a$, and $\alpha_s$ follow a three regime schedule governed by the clean validation metric $m_c$ and the adversarial validation metric $m_a$. During warm up, the loss terms use their initial weights. Once $m_c$ reaches the clean floor, the robust phase reduces the clean loss weight, increases the buffered adversarial loss weight, and increases the SICA loss scale. This schedule places greater emphasis on adversarial detection and texture shortcut suppression after the required clean performance has been reached.

\begin{algorithm}[t]
\small
\SetAlgoLined
\DontPrintSemicolon

\KwIn{
Detector $f_\theta$,
clean training set $\mathcal{D}_{\mathrm{cle}}$,
adversarial render buffer $B_{\mathrm{adv}}$ with capacity $m$,
and texture-snapshot queue $\mathcal{Q}_p$ with capacity $k$
}
\KwOut{Robust detector $f_\theta$}
\BlankLine

Initialise the online texture
$\mathbf{p}\in[0,1]^{3\times H_p\times W_p}$\;

Initialise
$B_{\mathrm{adv}}\leftarrow\varnothing$
and
$\mathcal{Q}_p\leftarrow\varnothing$\;

Prefill $B_{\mathrm{adv}}$ with $m$ adversarial renderings
and their detection labels\;

$m_c^0\leftarrow\mathrm{EvaluateClean}(f_\theta)$\;
$\phi_c\leftarrow m_c^0-\epsilon_c$\;
$m_a^\star\leftarrow0$\;

\For{epoch $=1$ \KwTo $E$}{

  $m_c\leftarrow\mathrm{EvaluateClean}(f_\theta)$\;

  $m_a\leftarrow
  \mathrm{EvaluateAdversarial}(f_\theta,\mathbf{p})$\;

  $(w_c,w_a,\alpha_s)\leftarrow
  \mathrm{ScheduleWeights}
  (m_c,\phi_c,m_a,m_a^\star,\epsilon_a)$\;

  $m_a^\star\leftarrow\max(m_a^\star,m_a)$\;

  \For{batch $(x,y)$ with index $b$}{

    $(x_{\mathrm{adv}},y_{\mathrm{adv}})
    \leftarrow
    \mathrm{Sample}
    (B_{\mathrm{adv}},\rho_{\mathrm{adv}}|x|)$\;

    $\mathcal{L}_{\mathrm{refresh}}\leftarrow0$\;

    \If{$b\bmod f_i=0$}{

      $(\mathcal{S},P_c,y_r,g_r)
      \leftarrow
      \mathrm{RenderClean}(n_r)$
      \tcp*{matched clean renderings and instance boxes}

      $\mathcal{Q}_p\leftarrow
      \mathrm{Push}
      (\mathcal{Q}_p,\mathrm{Detach}(\mathbf{p}))$\;

      $\mathbf{p}\leftarrow
      \mathrm{AdamPatchUpdate}
      (\mathbf{p},\mathcal{S},n_p;f_\theta)$
      \tcp*{freeze detector parameters}

      $P_a\leftarrow
      \mathrm{ReRender}(\mathbf{p},\mathcal{S})$\;

      $\widetilde{\mathbf{p}}\leftarrow
      \mathrm{SampleTexture}
      (\mathbf{p},\mathcal{Q}_p,p_s)$\;

      $x_{\mathrm{neg}}\leftarrow
      \mathrm{MakeNegative}
      (\widetilde{\mathbf{p}},g_r)$
      \tcp*{texture on a blank background}

      $\mathcal{L}_{\mathrm{refresh}}
      \leftarrow
      \mathcal{L}_{\mathrm{det}}(P_a,y_r)
      +
      \alpha_s
      \mathcal{L}_{\mathrm{SICA}}
      (P_a,P_c,x_{\mathrm{neg}},g_r)$\;

      $B_{\mathrm{adv}}\leftarrow
      \mathrm{Push}
      (B_{\mathrm{adv}},(P_a,y_r))$\;
    }

    $\mathcal{L}_{\mathrm{total}}
    \leftarrow
    w_c\mathcal{L}_{\mathrm{det}}(x,y)
    +
    w_a\mathcal{L}_{\mathrm{det}}
    (x_{\mathrm{adv}},y_{\mathrm{adv}})
    +
    \mathcal{L}_{\mathrm{refresh}}$\;

    Update $\Theta_{\mathrm{train}}$ using
    $\nabla_{\Theta_{\mathrm{train}}}
    \mathcal{L}_{\mathrm{total}}$\;
  }
}

\Return $f_\theta$\;

\caption{\inscat{} training}
\label{alg:inscat}
\end{algorithm}

\textbf{Adversarial guard.}
Long-horizon training can still suffer from robustness forgetting. As clean accuracy recovers in the late phase, adversarial validation performance under the current online texture can drop from its historical best . To prevent this degradation, an adversarial guard applies a still stronger setting when $m_a$ falls below its running best $m_a^\star$ by more than $\epsilon_a$ while clean is at the floor. Upon activation the guard lowers the clean loss weight $w_c$, raises the buffered adversarial loss weight $w_a$, and raises the \sica{} loss scale $\alpha_s$, with $w_c$ reduced to at most its guard-phase threshold so that clean performance is not abandoned. When $m_a$ recovers, the scheduler restores the normal robust-phase weights. The guard thus keeps the model on the robust trajectory that \ropo{} and \sica{} have reached, and it adds no new contrastive term.

Algorithm~\ref{alg:inscat} assembles the components introduced above into the complete \inscat{} training loop and specifies their execution order. At the beginning of each epoch, the gated scheduler determines the weights of the clean detection, buffered adversarial detection, and \sica{} objectives according to the clean and adversarial validation metrics. Every $f_i$ batches, \ropo{} samples a set of scene conditions, renders the corresponding clean instances, and updates the single global texture through inner maximisation with the detector frozen. The updated texture is then re-rendered under the same scene conditions to obtain matched adversarial instances. A texture-only negative is generated on a blank background using either the current texture or a snapshot sampled from its online optimisation trajectory. \sica{} applies the gap-aware loss to the resulting instance-level triplets, while the newly rendered adversarial samples and their detection labels are added to the adversarial render buffer. Between two refreshes, the detector continues to optimise the clean and buffered adversarial detection losses, thereby amortising the dominant rendering cost across multiple training batches.

\section{Experiments}
\label{sec:experiments}


\subsection{Experimental Setup}
\label{sec:experiments:setup}

\textbf{Datasets.}
Three data sources are used in this work.
1) MS-COCO 2017 is used for detector training~\cite{lin2014microsoft}. The detector is trained on the 118,287 images in train2017 over all 80 object categories. During training, a proportion $\rho_\text{adv}$ of the COCO images is augmented by pasting the current adversarial texture onto person regions. 2) The INRIAPerson dataset is used with strictly disjoint training and evaluation subsets~\cite{inriaperson}. The 614 positive images in Train-set are used exclusively by the auxiliary 2D attack branch during online texture optimisation. The 288 positive images in Test-set are reserved exclusively for digital paste attack evaluation and are never accessed during training. At test time, independently generated evaluation textures are pasted onto persons in the Test-set images to assess robustness in real-image person detection scenarios. 3) The nuScenes-mini dataset provides backgrounds for rendered adversarial samples~\cite{caesar2020nuscenes}. During training, adversarial human instances generated by the rendering pipeline are additionally composited onto 404 CAMFRONT background images. For rendered attack evaluation, adversarial human instances are composited onto 2,020 background images collected from the remaining five camera views, which are excluded from the training background set.

\textbf{Training and Evaluation Attacks.} During adversarial training, \ropo{} optimises a newly initialised global tiled clothing texture through joint 2D and 3D attack optimisation~\citep{huang2026advreal}. The training texture is generated entirely within the online optimisation process and is independent of all textures used for evaluation. At test time, the trained detector is evaluated on clean images, two non optimised control perturbations, and eight independently generated physical attack textures. RandomNoise and GrayBlock represent low level corruption and coarse occlusion. The adversarial effect focused attacks include AdvPatch~\citep{tom2017adversarial}, AdvTexture~\citep{hu2022adversarial}, AdvT-shirt~\citep{xu2020adversarial}, T-SEA~\citep{huang2023t}, AdvReal~\citep{huang2026advreal}, and AdvSerial~\citep{huang2026advserial}. The naturalness constrained attacks include NatPatch~\citep{hu2021naturalistic} and TRDPatch~\citep{wang2025transferable}. None of the evaluation textures is used during training, allowing the experiments to assess robustness to unseen attack textures and cross attack generalisation.

\begin{table*}[t!]
\centering
\caption{Cross attack generalisation of YOLOv5n on the nuScenes rendered domain in person AP. The rendered clean column reports person AP on clean rendered pedestrians. The remaining columns report person AP under independently generated evaluation textures.}
\label{tab:nuscences}
\setlength{\tabcolsep}{4.5pt}
\renewcommand{\arraystretch}{1}
\resizebox{\textwidth}{!}{%
\begin{tabular}{ll | cc | cccccccc | c}
\toprule
\textbf{Category} & \textbf{Method}
& \textbf{Clean} & \textbf{Noise} & \textbf{AdvPatch}
& \textbf{AdvTexture} & \textbf{AdvT-shirt} & \textbf{NatPatch} & \textbf{TRDPatch} & \textbf{TSEA}
& \textbf{AdvReal} & \textbf{AdvSerial} & \textbf{Avg} \\
\midrule
-- & Original
& 85.9 & 72.6 & 35.4 & 31.8 & 30.6 & 58.7 & 55.9 & 28.6 & 32.9 & 26.1 & 35.9 \\
\midrule
\multirow{3}{*}{Purify}
& LGS
& \underline{87.4} & 75.9 & 40.1 & 34.4 & 34.7 & 60.9 & 55.3 & 34.8 & 39.3 & 32.8 & 40.4 \\
& Jedi
& 82.0 & 18.9 & 16.7 & 14.7 & 12.0 & 56.7 & 27.8 & 13.1 & 21.2 & 18.3 & 23.3 \\
& PAD
& 80.6 & \underline{80.6} & \underline{74.9} & \underline{62.1} & \underline{79.8} & \underline{81.2} & \underline{76.2} & \underline{46.5} & \underline{56.1} & \underline{73.2} & \underline{71.2} \\
\midrule
\multirow{3}{*}{Detect}
& SAC
& 85.9 & 45.3 & 35.4 & 31.8 & 29.2 & 58.7 & 55.9 & 28.6 & 32.9 & 26.0 & 35.7 \\
& PatchZero
& 86.8 & 75.0 & 39.0 & 33.5 & 33.8 & 60.0 & 51.4 & 32.3 & 38.5 & 31.9 & 39.6 \\
& NAPGuard
& 72.4 & 57.1 & 21.2 & 25.1 & 24.7 & 25.3 & 38.3 & 19.4 & 30.3 & 11.8 & 23.1 \\
\midrule
\multirow{2}{*}{AT}
& $L_\infty$-AT (MTD)
& 83.5 & 70.8 & 38.6 & 33.2 & 32.1 & 59.4 & 57.3 & 31.9 & 35.8 & 28.7 & 37.2 \\

& \textbf{InsCAT (Ours)}
& \textbf{96.0} & \textbf{93.5} & \textbf{84.7} & \textbf{81.2}
& \textbf{84.6} & \textbf{89.3} & \textbf{84.2} & \textbf{74.8} & \textbf{73.7} & \textbf{80.3} & \textbf{82.3} \\
\bottomrule
\end{tabular}%
}
  \vspace{-4 mm} 
\end{table*}

\begin{table*}[t!]
\centering
\caption{Cross attack generalisation of YOLOv5n on INRIAPerson in person AP. All defence methods use the same YOLOv5n detector or are trained with the same YOLOv5n architecture.}
\label{tab:inria}
\renewcommand{\arraystretch}{1}
\setlength{\tabcolsep}{3pt}
\resizebox{\textwidth}{!}{
\begin{tabular}{ll|ccc|cccccccc|c}
\toprule
\textbf{Category} & \textbf{Method}
& \textbf{Clean} & \textbf{Noise} & \textbf{Gray}
& \textbf{AdvPatch} & \textbf{AdvTexture} & \textbf{AdvT-shirt}
& \textbf{NatPatch} & \textbf{TRDPatch} & \textbf{TSEA}
& \textbf{AdvReal} & \textbf{AdvSerial} & \textbf{Avg} \\
\midrule
-- & Original
& \underline{96.8} & 95.5 & 95.2 & 81.1 & 77.3 & 85.1 & 88.5 & 66.2 & 36.9 & 35.0 & 75.2 & 68.2 \\
\midrule
\multirow{3}{*}{Purify}
& LGS
& \textbf{96.9} & \underline{95.7} & \underline{95.4} & 84.3 & 84.8 & 89.0 & 91.7 & 69.6 & 36.4 & 34.9 & 79.7 & 71.3 \\
& Jedi
& 96.7 & 95.4 & 95.1 & 81.5 & 77.7 & 85.5 & 88.2 & 67.7 & 38.0 & 35.2 & 75.1 & 68.6 \\
& PAD
& 95.8 & 94.9 & 94.5 & \textbf{94.2} & \textbf{89.7} & \textbf{93.3}
& \textbf{93.4} & \textbf{95.7} & \textbf{82.7} & \textbf{72.2} & \textbf{93.0} & \textbf{89.3} \\
\midrule
\multirow{3}{*}{Detect}
& SAC
& \underline{96.8} & 95.1 & 95.2 & 80.8 & 76.6 & 85.2 & 88.4 & 65.8 & 37.4 & 34.9 & 75.0 & 68.0 \\
& PatchZero
& 96.6 & \textbf{95.8} & \textbf{95.5} & 83.6 & 80.2 & 87.9 & 90.8 & 68.1 & 38.5 & 36.1 & 78.9 & 70.2 \\
& NAPGuard
& \textbf{96.9} & 95.5 & 95.2 & 81.6 & 77.4 & 85.8 & 88.5 & 65.8 & 37.0 & 35.0 & 75.2 & 68.3 \\
\midrule
\multirow{2}{*}{AT}
& $L_\infty$-AT (MTD)
& 95.1 & 93.2 & 92.9 & 83.4 & 80.6 & 87.1 & 89.6 & 71.5 & 41.2 & 46.3 & 84.5 & 73.0 \\
& \textbf{InsCAT (Ours)}
& 95.9 & 94.3 & 94.1 & \underline{87.2} & \underline{86.6} & \underline{91.3}
& \underline{91.9} & \underline{79.2} & \underline{47.7} & \underline{54.7} & \underline{88.0} & \underline{78.3} \\
\bottomrule
\end{tabular}
}
  \vspace{-4 mm} 
\end{table*}

\textbf{Detectors.} The main experiments are conducted on YOLOv5n~\cite{yolov5}, a lightweight one-stage detector with a $640{\times}640$ input resolution. To verify that \inscat{} is not architecture-specific, we additionally apply it to Faster R-CNN ~\citep{ren2016fasterrcnnrealtimeobject} and DN-DETR ~\cite{li2022dn}, representing two-stage and Transformer-based paradigms respectively.

\textbf{Alignment layer.}
Following the design in Section~\ref{sec:method:layer}, \sica{} is attached to a late spatial feature map before the architecture-specific detection head. This placement preserves spatial correspondence for ground-truth-aligned RoI extraction while avoiding detector-specific prediction structures. For YOLOv5n, we use the P5 neck feature with stride 32, as selected by the layer study in Tab.~\ref{tab:layer_select}. For Faster R-CNN, \sica{} operates on the FPN-P5 neck feature map. For DN-DETR, we use the last backbone-stage feature map and apply RoIAlign with the ground-truth person boxes to extract instance-level representations. These attachment points provide a consistent instance representation interface across one-stage, two-stage, and Transformer-based detectors.

\textbf{Training.}
All detectors are trained for at most $100$ epochs with cosine learning rate decay and early stopping after $10$ epochs without validation improvement. The main experiments use an adversarial mix ratio of $\rho_{\mathrm{adv}}{=}0.30$. \ropo{} maintains a render buffer of $m{=}120$ samples and refreshes it every $f_i{=}20$ detector updates. Each refresh renders $n_r{=}4$ scene configurations and performs $n_p{=}5$ Adam steps on the shared texture. The tiled UV texture uses a scale of $s_t{=}0.45$, with total variation weight $\lambda_{\mathrm{tv}}{=}3.5$.

For \sica{}, the base loss weight is $\lambda{=}0.3$, the similarity gap is set to $\delta{=}0.35$, and the negative similarity margin is $m_{\mathrm{neg}}{=}0.10$. Trajectory snapshots are sampled with probability $p_s{=}0.70$, and SICA is attached to the P5 neck feature map layer selected in Tab.~\ref{tab:layer_select}. The clean and adversarial validation tolerances are both set to $\epsilon_c{=}\epsilon_a{=}0.03$. During robust training, the buffered adversarial loss weight is increased to $w_a{=}2.0$ and the SICA scale to $\alpha_s{=}1.5$, while the clean loss weight is maintained at $w_c{=}0.5$. Detector specific optimisers and learning rates follow their standard training configurations.

\textbf{Baselines and metrics.}
Adversarial-training baselines (\pbcat{}, \inscat{}) are retrained and differ only in defence; input-purification baselines (SAC, PatchZero, Jedi, PAD, NAPGuard, LGS) run as pre-processing in front of the \emph{same} frozen detector, so the comparison is not confounded by a different backbone. The primary metric is person AP; we additionally report the texture FPR on human-free patch texture as the operational measure of the shortcut, together with the attack success rate (ASR). AT-Mix is an internal adversarial-training baseline that uses the same \ropo{}-based online texture generation, adversarial sample mixing, and Guard schedule as \inscat{}, but removes the \sica{} objective. It therefore controls for adversarial exposure and training coordination without explicit feature-level alignment.

\subsection{Comparative Experiments}

\subsubsection{Defensive Capabilities}

We compare \inscat{} with seven representative defence methods on YOLOv5n, including the input purification methods LGS, Jedi, PAD, SAC, PatchZero, and NAPGuard, together with the adversarial training baseline $\ell_\infty$ AT. Tables~\ref{tab:nuscences} and~\ref{tab:inria} report cross attack performance on the nuScenes rendered domain and the INRIAPerson digital domain. PBCAT is examined separately in Section~\ref{sec:cross_detector} using its supported Faster R-CNN and DN-DETR architectures.

On the nuScenes rendered domain in Tab.~\ref{tab:nuscences}, \inscat{} achieves the highest rendered clean person AP and the highest average attack AP. The undefended Original obtains an average attack AP of $35.9$, while most existing defences remain between $23.1$ and $40.4$. PAD is the strongest baseline with an average attack AP of $71.2$, together with a rendered clean person AP of $80.6$. \inscat{} reaches $82.3$ average attack AP and $96.0$ rendered clean person AP, outperforming PAD by 11.1 points in average attack AP. The gain over PAD is observed across all eight evaluation textures and is particularly large on AdvTexture, TSEA, and AdvReal. These results show that \inscat{} achieves a more favourable balance between clean and adversarial person detection within the rendered domain.


Tab.~\ref{tab:inria} reports person AP on the INRIAPerson digital paste domain under eight independently generated attacks, none of which is used during \inscat{} training. PAD achieves the highest average attack AP of $89.3$ and leads on all eight attacks. \inscat{} ranks second with an average AP of $78.3$, exceeding the strongest adversarial training baseline, $L_\infty$-AT, by $5.3$ points. The largest gaps between PAD and \inscat{} occur on TSEA, AdvReal, and TRDPatch, where PAD leads by $35.0$, $17.5$, and $16.5$ points. On the remaining five attacks, the gap is at most $7.0$ points, including $2.0$ points on AdvT-shirt and $1.5$ points on NatPatch. The two methods retain nearly identical clean accuracy, with $95.8$ for PAD and $95.9$ for \inscat{}.

INRIAPerson evaluates attacks through static 2D composition and does not reproduce garment deformation, viewpoint changes, illumination variation, or temporal motion. This setting is particularly favourable to preprocessing defences that analyse and modify each image before detection. PAD therefore provides a strong robustness reference in the digital domain, while \inscat{} remains the strongest training based defence. Since preprocessing can introduce substantial deployment overhead, the next section compares the computational latency of all defence methods.


\subsubsection{Computing Latency}

\begin{table}[h]
\centering
\caption{Per-image inference latency (ms) of each defense on YOLOv5n with Nvidia RTX 4080, split into preprocessing and inference. The parenthesized value is the overhead over the Undefended baseline.}
\label{tab:latency}
\renewcommand{\arraystretch}{1.15}
\setlength{\tabcolsep}{6pt}
\resizebox{8.6cm}{!}{
\begin{tabular}{lccl}
\toprule
\multirow{2}{*}{\textbf{Method}} & \multicolumn{3}{c}{\textbf{Latency (ms/img)}} \\
\cmidrule(lr){2-4}
 & Preprocess & Inference & \multicolumn{1}{c}{Total (overhead)} \\
\midrule
Undefended             & --      & 0.254 & 0.25\,{\scriptsize(+0.00)}       \\
LGS                    & 70.01   & 0.542 & 70.55\,{\scriptsize(+70.30)}     \\
Jedi                   & 553.83  & 0.255 & 554.09\,{\scriptsize(+553.84)}   \\
PAD                    & 8601.25 & 0.274 & 8601.52\,{\scriptsize(+8601.27)} \\
SAC                    & 111.82  & 0.249 & 112.07\,{\scriptsize(+111.82)}   \\
PatchZero              & 220.07  & 0.257 & 220.33\,{\scriptsize(+220.08)}   \\
NAPGuard-Occlude       & 99.68   & 0.257 & 99.94\,{\scriptsize(+99.69)}     \\
$\ell_\infty$-AT (MTD) & --      & 0.273 & \textbf{0.27\,{\scriptsize(+0.02)}}       \\
\rowcolor{lightgrayrow} \textbf{InsCAT (ours)} & --      & 0.285 & \underline{0.29\,{\scriptsize(+0.04)}}       \\
\bottomrule
\end{tabular}
}
\vspace{-4 mm} 
\end{table}

\begin{figure}[t]
  \centering
  \includegraphics[width=8cm]{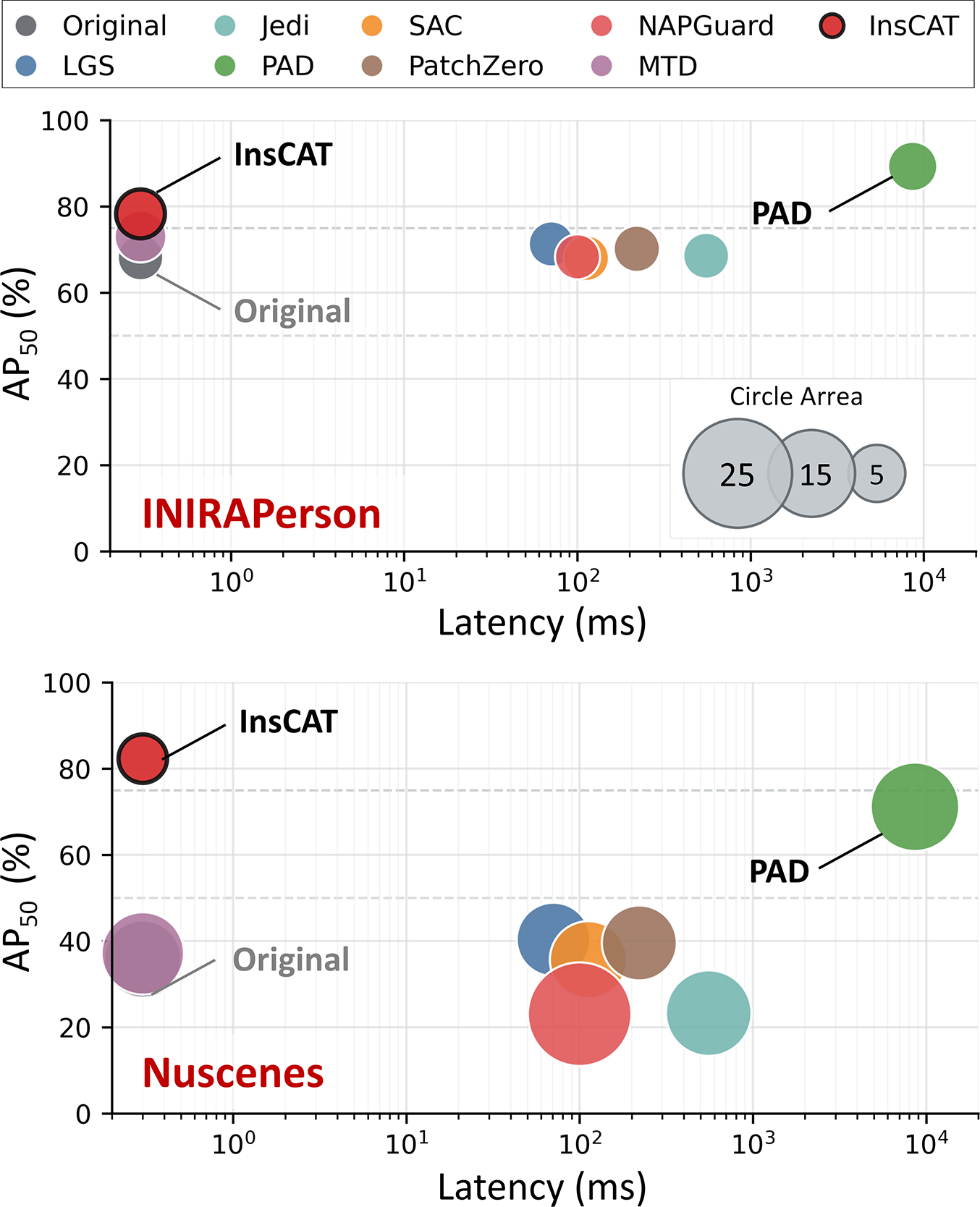}
  \caption{Performance–latency trade-off comparison of different methods on INRIAPerson and NuScenes.}
  \label{fig:latency}
  \vspace{-4 mm} 
\end{figure}

However, the advantage of PAD comes at a deployment cost that disqualifies it from any online use. Tab.~\ref{tab:latency} splits per-frame latency into preprocessing and inference. PAD spends 8601 ms per image, dominated entirely by its preprocessing stage, which corresponds to roughly 0.12 frames per second and falls far below any real-time budget. The other preprocess-and-detect methods also pay a large preprocessing cost, from 70 ms for LGS up to 554 ms for Jedi. \inscat{} operates as a direct detector with no preprocessing stage, so its total latency is 0.29 ms per image, an overhead of only 0.04 ms over the undefended baseline and on par with $\ell_\infty$-AT (MTD) at 0.27 ms. Fig.~\ref{fig:latency} makes the trade-off explicit on both INRIAPerson and nuScenes. The x-axis is per-frame latency on a log scale spanning four orders of magnitude, and the y-axis is AP$_{50}$ under attack. InsCAT lies in the upper-left region of both panels, combining high robustness with sub-millisecond latency. PAD sits at the upper-right corner, reaching comparable or higher AP$_{50}$ but at a latency nearly four orders of magnitude larger. The remaining preprocess-and-detect methods are scattered in between and are dominated by \inscat{}, since they are slower yet less robust. \inscat{} is the only method that combines strong cross-attack robustness with real-time deployment cost, whereas PAD, despite its leading INRIAPerson robustness, is impractical for any online detector.

\subsubsection{ROPO Training Efficiency}
\label{sec:ropo_efficiency}

\begin{table}[h]
\centering
\caption{Training efficiency of ROPO. FreshRender generates new adversarial renderings for every detector batch, while ROPO reuses samples from the render buffer.}
\label{tab:ropo_efficiency}
\renewcommand{\arraystretch}{1.15}
\setlength{\tabcolsep}{4.5pt}
\resizebox{7cm}{!}{
\begin{tabular}{lcccc}
\toprule
\textbf{Method} & $f_i$ & \textbf{Render calls} & \textbf{Time (s)} & \textbf{Throughput} \\
\midrule
FreshRender & 20 & 4907 & 427.2 & 37.5 \\
ROPO & 10 & 345 & 44.6 & 359.0 \\
ROPO & 20 & 176 & \textbf{40.9} & \textbf{391.5} \\
\bottomrule
\end{tabular}}
\vspace{-4 mm}
\end{table}

Differentiable rendering is the dominant computational cost in the InsCAT training pipeline. ROPO separates texture refresh from detector optimisation through a render buffer. The adversarial texture and rendered samples are refreshed every $f_i$ detector batches, while the intervening batches reuse buffered adversarial samples.

We compare ROPO with FreshRender, which follows the same detector training schedule and updates the adversarial texture every $20$ batches but renders new adversarial samples for every detector batch. Both methods use the same initial weights, random seed, scene sequence, and batch size of $64$. Each run contains $300$ detector batches on a single RTX 4080. The first $50$ batches are used for warm up and the remaining $250$ batches are timed. SICA and Guard are disabled to isolate the cost of rendering reuse.

FreshRender invokes PyTorch3D $4907$ times during the timed interval, including $4751$ per batch renderings and $156$ renderings used for texture updates. ROPO with $f_i{=}20$ reduces the number of rendering calls to $176$, a $27.9\times$ reduction, and shortens the total training time from $427.2$ s to $40.9$ s. This corresponds to a $10.45\times$ end to end speedup and increases throughput from $37.5$ to $391.5$ samples per second. Reducing the refresh interval from $20$ to $10$ increases the number of rendering calls from $176$ to $345$, while the total time changes only from $40.9$ s to $44.6$ s. We use $f_i{=}20$ to retain regularly refreshed adversarial samples with substantially lower rendering cost.

\subsection{Ablation Studies and Shortcut Learning}
\label{sec:results:ablation}

\subsubsection{Module Ablation}

\begin{table*}[t]
\centering
\caption{Component ablation of InsCAT. Configurations A to C progressively add Pull and Push to ROPO. Configuration D combines ROPO with Guard, and Configuration E denotes the full model. The similarity gap is defined as $s_{\mathrm{pos}}-s_{\mathrm{neg}}$. Higher AP and similarity gap are better, while lower $s_{\mathrm{neg}}$ is better. The best and second best results are marked in \textbf{bold} and \underline{underlined}, respectively.}
\label{tab:ablation}
\renewcommand{\arraystretch}{1.15}
\setlength{\tabcolsep}{5pt}
\resizebox{14cm}{!}{
\begin{tabular}{lcccccccc}
\toprule
\multirow{2}{*}{\textbf{Config}}
& \multicolumn{4}{c}{\textbf{Component}}
& \multicolumn{4}{c}{\textbf{Detection and Feature Statistics}} \\
\cmidrule(lr){2-5}\cmidrule(lr){6-9}
& \textbf{ROPO} & \textbf{Pull} & \textbf{Push} & \textbf{Guard}
& \textbf{adv AP$_{50}\uparrow$}
& \textbf{adv AP$_{50{:}95}\uparrow$}
& $s_{\mathrm{neg}}\downarrow$
& \textbf{Similarity gap$\uparrow$} \\
\midrule
A ROPO only
& \checkmark & -- & -- & --
& 0.731 & 0.442 & 0.317 & 0.119 \\

B +SICA Pull
& \checkmark & \checkmark & -- & --
& 0.734 & 0.449 & \underline{0.303} & \underline{0.200} \\

C +SICA Pull Push
& \checkmark & \checkmark & \checkmark & --
& 0.739 & 0.451 & 0.328 & 0.016 \\

D AT-Mix
& \checkmark & -- & -- & \checkmark
& \textbf{0.802} & \textbf{0.529} & 0.333 & 0.124 \\

E Full InsCAT
& \checkmark & \checkmark & \checkmark & \checkmark
& \underline{0.773} & \underline{0.486}
& \textbf{0.140} & \textbf{0.239} \\
\bottomrule
\end{tabular}
}
\vspace{-4 mm}
\end{table*}

The ablation study reports detection performance and online feature statistics for the five configurations in Tab.~\ref{tab:ablation}. Adversarial AP$_{50}$ and AP$_{50{:}95}$ are evaluated for the person class on nuScenes composites generated with a fixed adversarial texture. The texture similarity $s_{\mathrm{neg}}$ measures the association between adversarial person features and texture negative features, while the similarity gap $s_{\mathrm{pos}}-s_{\mathrm{neg}}$ measures their separation relative to the corresponding clean person features.

Pull increases the similarity gap from $0.119$ to $0.200$ and reduces $s_{\mathrm{neg}}$ from $0.317$ to $0.303$, while adversarial AP$_{50}$ changes only from $0.731$ to $0.734$. Positive alignment therefore improves the relative ordering of clean and texture representations before producing a substantial detection gain. Adding Push without Guard raises adversarial AP$_{50}$ to $0.739$, but contracts the similarity gap to $0.016$ and increases $s_{\mathrm{neg}}$ to $0.328$. Under this training condition, the detection gain is accompanied by a deterioration in the intended feature separation, showing that the Push objective requires coordinated optimisation rather than isolated strengthening.

Guard enables this coordination in the full model. Compared with Configuration C, full InsCAT increases adversarial AP$_{50}$ from $0.739$ to $0.773$, reduces $s_{\mathrm{neg}}$ from $0.328$ to $0.140$, and expands the similarity gap from $0.016$ to $0.239$. The comparison with AT-Mix further exposes the texture shortcut. AT-Mix achieves the highest adversarial AP$_{50}$ of $0.802$, while retaining the highest $s_{\mathrm{neg}}$ of $0.333$ and a similarity gap of only $0.124$. Full InsCAT reduces texture similarity by $58.0\%$ and nearly doubles the similarity gap relative to AT-Mix, with adversarial AP$_{50}$ decreasing by $2.9$ points. These results show that strong performance against the current adversarial texture can still rely on texture related features. SICA and Guard jointly improve the feature separation associated with shortcut resistance while maintaining high adversarial detection performance.

\subsubsection{Texture Shortcut Diagnosis}

\begin{figure}[h!]
  \centering
  \includegraphics[width=8.6cm]{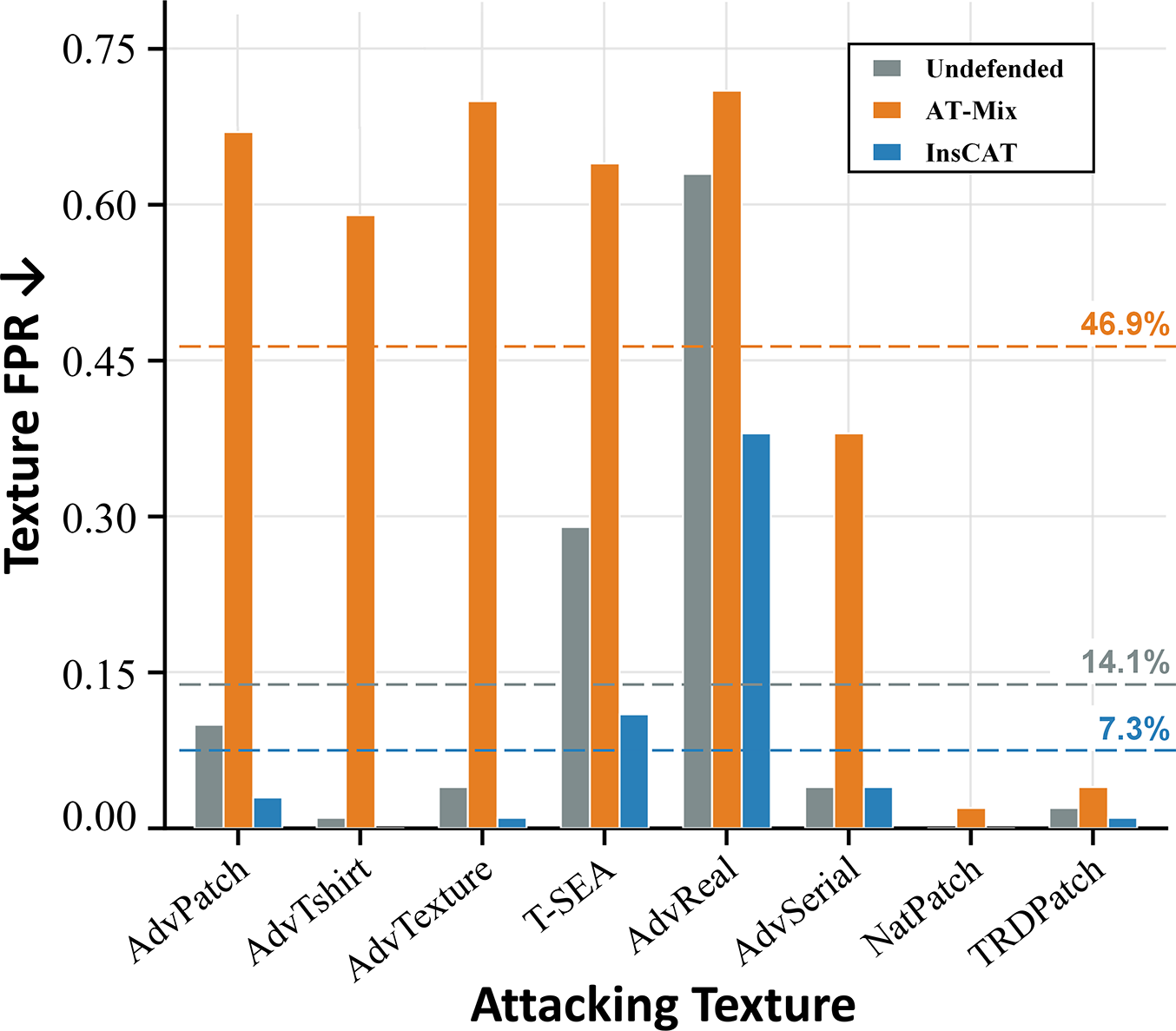}
  \caption{Texture false-positive rate on human-free texture-only images. AT-Mix shows severe texture-triggered person hallucination, whereas InsCAT maintains low texture FPR while preserving cross-attack robustness. Dashed lines denote the average FPR of each method.}
  \label{fig:texture_shortcut}
  \vspace{-4 mm} 
\end{figure}

The ablation results show that AT-Mix achieves strong adversarial detection performance while maintaining a relatively high similarity between adversarial and texture-only features. This motivates a further diagnosis of texture shortcut learning. We evaluate each model on human-free texture-only images, where each attack texture is presented without any human instance and every detected person is counted as a false positive. A robust detector should maintain a low texture false-positive rate because the texture alone should not provide sufficient evidence for the person category.

Fig.~\ref{fig:texture_shortcut} compares the texture FPR of the undefended model, AT-Mix, and InsCAT across eight attack textures. AT-Mix reaches an average texture FPR of 46.9\%, which is more than three times that of the undefended model at 14.1\%. Particularly high false-positive rates are observed on AdvPatch, AdvTexture, AdvTshirt, T-SEA, and AdvReal, reaching 67\%, 70\%, 59\%, 64\%, and 71\%, respectively. These results show that adversarial training without explicit shortcut suppression strengthens the association between adversarial texture and the person category, causing the detector to respond to attack textures even in the absence of human structure.

InsCAT reduces the average texture FPR to 7.3\%, which is approximately one sixth of that of AT-Mix and is also lower than the undefended model. Across all eight attack textures, the FPR of InsCAT never exceeds that of the undefended detector and is reduced on six of them. Although AT-Mix achieves an adversarial AP of 78.1, approaching the 82.3 achieved by InsCAT, its texture FPR is 6.4 times higher. This result shows that adversarial AP alone does not fully characterize robust detection because improved detection performance can be accompanied by severe texture-triggered false positives. InsCAT maintains high adversarial AP while substantially reducing texture-induced false detections, providing a more reliable robustness profile with lower dependence on texture-only cues.

\subsection{Sensitivity Study}
\label{sec:sensitive}


\textbf{Feature Layer Selection.}
Tab.~\ref{tab:layer_select} compares P3, P4, and P5 as candidate SICA attachment layers. We define $s_{\mathrm{pos}}=\cos(a,c)$, $s_{\mathrm{neg}}=\cos(a,t)$, $\mathrm{gap}=s_{\mathrm{pos}}-s_{\mathrm{neg}}$, and $d_{\mathrm{neg}}=1-s_{\mathrm{neg}}$, where $a$, $c$, and $t$ denote adversarial person, matched clean person, and texture only representations. The Undefended and Full InsCAT columns measure these feature relationships at each layer using the corresponding trained model. The Layer-specific SICA columns report three separate training runs with SICA attached to P3, P4, or P5.

\begin{table}[h]
\centering
\caption{Feature diagnostics for candidate SICA layers. The first two groups report features measured at each layer in the undefended and full InsCAT models. Layer-specific SICA reports separate training runs with SICA attached to P3, P4, or P5. P5 is selected and shaded.}
\label{tab:layer_select}
\renewcommand{\arraystretch}{1.22}
\setlength{\tabcolsep}{4.2pt}
\resizebox{8.6cm}{!}{
\begin{tabular}{lc|cc|cc|cc}
\toprule
\multirow{2}{*}{\textbf{Layer}} &
\multirow{2}{*}{\textbf{Res.}} &
\multicolumn{2}{c|}{\textit{Undefended}} &
\multicolumn{2}{c|}{\textit{Full InsCAT}} &
\multicolumn{2}{c}{\textit{Layer-specific SICA}} \\
\cmidrule(lr){3-4}\cmidrule(lr){5-6}\cmidrule(lr){7-8}
& & $s_{\mathrm{neg}}\!\downarrow$ & gap$\uparrow$
& $s_{\mathrm{neg}}\!\downarrow$ & gap$\uparrow$
& gap$\uparrow$ & $d_{\mathrm{neg}}\!\uparrow$ \\
\midrule
P3 & $80{\times}80$ & 0.558 & 0.344 & 0.693 & 0.252 & 0.328 & 0.410 \\
P4 & $40{\times}40$ & 0.408 & \textbf{0.448} & 0.550 & 0.363 & 0.428 & 0.556 \\
\rowcolor{lightgrayrow}
\textbf{P5} & $20{\times}20$ & \textbf{0.375} & 0.446 & \textbf{0.442} & \textbf{0.449} & \textbf{0.447} & \textbf{0.625} \\
\bottomrule
\end{tabular}
}
\vspace{-4 mm}
\end{table}

P3 produces the weakest feature separation, with the highest $s_{\mathrm{neg}}$ and the smallest similarity gap after InsCAT training. P4 improves the separation substantially, while P5 achieves the strongest overall result. Under Full InsCAT, P5 reaches the lowest $s_{\mathrm{neg}}$ of $0.442$ and the highest similarity gap of $0.449$. In the layer-specific training runs, P5 also obtains the largest gap of $0.447$ and the highest $d_{\mathrm{neg}}$ of $0.625$. These results show that the deeper P5 representation provides the clearest separation between adversarial person and texture only features. We therefore select P5 as the default SICA attachment layer.


\textbf{Hyper-parameter Sensitivity.}
We analyse the global SICA loss scale $\lambda$ and the negative separation weight $w_{\mathrm{neg}}$. The former controls the overall contribution of the contrastive objective, while the latter weights the penalty applied when adversarial person features remain close to texture only features. The remaining internal weights are fixed at $w_{\mathrm{pos}}{=}0.05$ and $w_{\mathrm{gap}}{=}0.8$. Tab.~\ref{tab:sensitivity} reports the mean and standard deviation of clean and adversarial person AP$_{50}$ on INRIAPerson, together with average adversarial AP$_{50}$ and AP$_{50{:}95}$ under eight independently generated attack textures in the nuScenes rendered domain.

\begin{table}[t]
\centering
\caption{Sensitivity to SICA hyperparameters. Shaded rows denote the default settings, and bold values mark the highest mean in each sweep.}
\label{tab:sensitivity}

\renewcommand{\arraystretch}{1.18}
\setlength{\tabcolsep}{9pt}

\newcommand{\ms}[2]{$#1_{\scriptscriptstyle \pm #2}$}
\newcommand{\bms}[2]{$\mathbf{#1}_{\scriptscriptstyle \pm #2}$}

\resizebox{8.6cm}{!}{
\begin{tabular}{cccccc}
\toprule
\multirow{2}{*}{\textbf{Para.}}
& \multirow{2}{*}{\textbf{Value}}
& \multicolumn{2}{c}{\textbf{INRIAPerson}}
& \multicolumn{2}{c}{\textbf{nuScenes}} \\
\cmidrule(lr){3-4}
\cmidrule(lr){5-6}
&
& \textbf{Clean$\uparrow$}
& \textbf{Adv.$\uparrow$}
& \textbf{AP$_{50}\uparrow$}
& \textbf{AP$_{50{:}95}\uparrow$} \\
\midrule

$\lambda$
& 0.1
& \ms{0.942}{0.004}
& \ms{0.488}{0.014}
& \bms{0.806}{0.026}
& \bms{0.511}{0.021} \\

\rowcolor{lightgrayrow}
& 0.3
& \ms{0.941}{0.003}
& \ms{0.488}{0.014}
& \ms{0.804}{0.028}
& \ms{0.508}{0.027} \\

& 0.5
& \bms{0.943}{0.003}
& \bms{0.493}{0.015}
& \ms{0.803}{0.027}
& \ms{0.509}{0.022} \\

\midrule

$w_{\mathrm{neg}}$
& 0.4
& \bms{0.943}{0.004}
& \bms{0.495}{0.011}
& \ms{0.806}{0.029}
& \ms{0.512}{0.028} \\

\rowcolor{lightgrayrow}
& 0.8
& \ms{0.941}{0.003}
& \ms{0.488}{0.014}
& \ms{0.804}{0.028}
& \ms{0.508}{0.027} \\

& 1.2
& \ms{0.941}{0.004}
& \ms{0.492}{0.015}
& \bms{0.811}{0.028}
& \bms{0.516}{0.027} \\

\bottomrule
\end{tabular}
}

\vspace{-4mm}
\end{table}

Performance remains stable across the evaluated values of $\lambda$. INRIA clean AP$_{50}$ varies by at most $0.2$ points, INRIA adversarial AP$_{50}$ by $0.5$ points, and both nuScenes metrics by at most $0.3$ points. These differences are substantially smaller than the corresponding run to run standard deviations, showing that the overall strength of the SICA objective does not require precise tuning.

The effect of $w_{\mathrm{neg}}$ is similarly limited. A value of $0.4$ obtains the highest mean performance on INRIAPerson, while $1.2$ obtains the highest means on nuScenes. The largest difference across this sweep is only $0.8$ points and remains smaller than the observed standard deviations. No single value therefore provides a consistent advantage across both evaluation domains. We retain $\lambda{=}0.3$ and $w_{\mathrm{neg}}{=}0.8$ as central default settings throughout the main experiments. Overall, SICA is stable across the evaluated parameter ranges and does not depend on delicate hyperparameter selection.

\begin{figure*}[t]
  \centering
  \includegraphics[width=18cm]{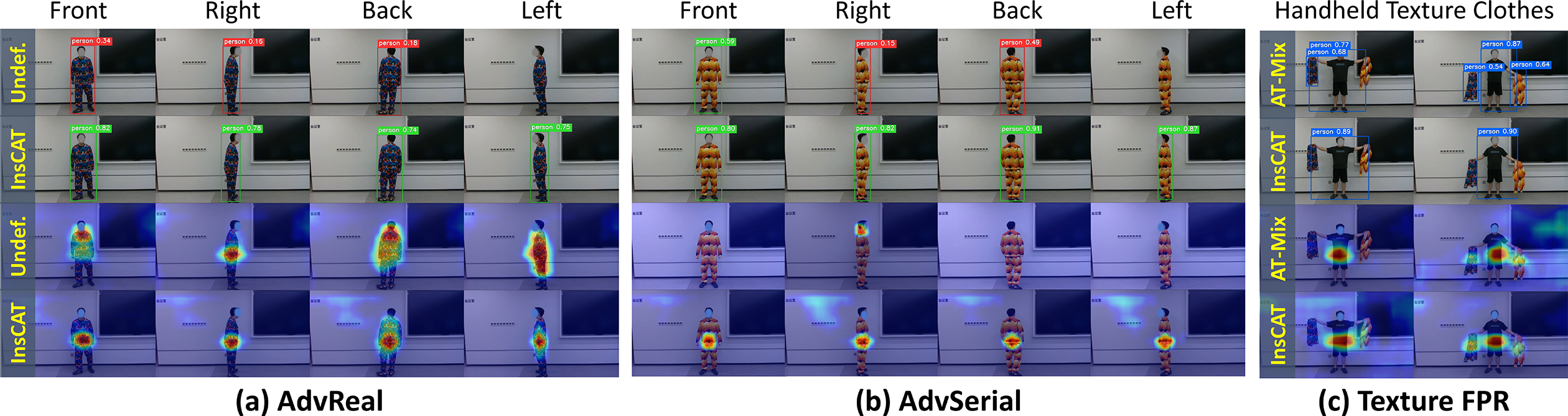}
  \caption{Physical-world detection results and Grad-CAM visualizations under AdvReal and AdvSerial. InsCAT improves person detection while reducing responses to adversarial textures.}
  \label{fig:physical}
  \vspace{-4 mm} 
\end{figure*}


\subsection{Cross-detector evaluation. }
\label{sec:cross_detector}

To examine whether the learned robustness is tied to a specific detector architecture or evaluation domain, we further evaluate InsCAT on YOLOv5n, Faster R-CNN, and DN-DETR, representing one stage, two stage, and Transformer based detection paradigms. Tab.~\ref{tab:cross_detector} reports clean and adversarial person AP$_{50}$ on INRIAPerson and the nuScenes rendered domain. INRIAPerson evaluates digitally composed attack textures on real images, while nuScenes introduces three dimensional rendering variations in driving scenes. PBCAT is additionally evaluated on Faster R-CNN and DN-DETR under the same protocols.

\begin{table}[h]
\centering
\caption{Cross-detector evaluation on INRIAPerson and the nuScenes rendered domain. Adv denotes the average person AP$_{50}$ under eight independently generated attack textures.}
\label{tab:cross_detector}
\renewcommand{\arraystretch}{1.12}
\setlength{\tabcolsep}{3.8pt}
\resizebox{7cm}{!}{
\begin{tabular}{llcc|cc}
\toprule
\multirow{2}{*}{\textbf{Detector}}
& \multirow{2}{*}{\textbf{Method}}
& \multicolumn{2}{c|}{\textbf{INRIAPerson}}
& \multicolumn{2}{c}{\textbf{nuScenes}} \\
\cmidrule(lr){3-4}
\cmidrule(lr){5-6}
&
& \textbf{Clean$\uparrow$}
& \textbf{Adv$\uparrow$}
& \textbf{Clean$\uparrow$}
& \textbf{Adv$\uparrow$} \\
\midrule

\multirow{2}{*}{YOLOv5n}
& Pretrained
& \textbf{96.8} & 68.2
& 85.9 & 35.9 \\
& InsCAT
& 95.9 & \textbf{78.3}
& \textbf{96.0} & \textbf{82.3} \\

\midrule
\multirow{3}{*}{F-RCNN}
& Pretrained
& 98.5 & 86.2
& 77.9 & 52.4 \\
& PBCAT
& 97.7 & \underline{91.3}
& \underline{86.8} & \underline{66.5} \\
& InsCAT
& \textbf{98.6} & \textbf{97.3}
& \textbf{89.4} & \textbf{76.2} \\

\midrule
\multirow{3}{*}{DN-DETR}
& Pretrained
& \textbf{93.7} & 59.2
& 77.5 & 34.0 \\
& PBCAT
& 92.9 & \underline{79.1}
& \underline{85.6} & \underline{60.8} \\
& InsCAT
& \underline{93.6} & \textbf{85.8}
& \textbf{88.7} & \textbf{72.5} \\

\bottomrule
\end{tabular}
}
\vspace{-4mm}
\end{table}

InsCAT consistently improves adversarial detection across all three architectures and both evaluation domains. Relative to the pretrained detectors, the gains range from $10.1$ to $26.6$ points on INRIAPerson and from $23.8$ to $46.4$ points on nuScenes. The improvements are observed for anchor based prediction, region proposal based detection, and object query based detection. They also persist across digital composition and three dimensional rendering, indicating that the learned robustness is not specific to one detector family or attack presentation.

The comparison with PBCAT further separates feature regulation from attack exposure alone. On Faster R-CNN and DN-DETR, InsCAT improves adversarial AP$_{50}$ over PBCAT by $6.0$ and $6.7$ points on INRIAPerson, and by $9.7$ and $11.7$ points on nuScenes. The consistent advantage in both domains shows that expanding the adversarial training distribution does not fully address the texture dependence induced during training. Explicitly aligning adversarial person features with their clean counterparts and separating them from texture only features provides additional robustness across architectures and physical variations.

Clean person detection is preserved on INRIAPerson, where the change relative to the pretrained models remains within $0.9$ points. On nuScenes, InsCAT also improves clean rendered AP by more than $10$ points across all three detectors. The robustness gains therefore do not result from sacrificing normal person detection. Together with the texture false positive and feature similarity analyses, these results support that InsCAT improves adversarial detection by reducing reliance on texture related evidence while retaining target relevant representations.

\subsection{Physical-World Evaluation}
\label{sec:results:physical}

\textbf{Physical-World performance.} To evaluate the transferability of digital adversarial patterns to physical environments, we conduct real-world experiments using adversarial clothes generated by AdvReal-v5 and AdvSerial-v5. Two video sequences are captured under different human poses and motion conditions, including posture changes and continuous movements. The video sequences contain 6372 frames and 6372 ground-truth pedestrian instances in total, with 3186 frames captured for each adversarial garment. All videos are recorded using the same camera setup with Intel D435i and evaluated directly without additional adaptation, providing a practical assessment of the robustness of InsCAT against physical adversarial appearances.

\begin{table}[t]
\centering
\caption{Physical-world evaluation on captured videos. ASR and FPR denote the attack success rate and false positive rate, respectively. The best results are highlighted in \textbf{bold}.}
\label{tab:physical}
\renewcommand{\arraystretch}{1.12}
\setlength{\tabcolsep}{4pt}
\resizebox{8cm}{!}{
\begin{tabular}{lccccc}
\toprule
\textbf{Model} 
& \textbf{Precision}$\uparrow$ 
& \textbf{Recall}$\uparrow$ 
& \textbf{F1}$\uparrow$ 
& \textbf{ASR}$\downarrow$ 
& \textbf{FPR}$\downarrow$ \\
\midrule
Undefended 
& 97.8\% & 78.2\% & 86.9\% & 21.8\% & 2.2\% \\
AT-Mix 
& 67.5\% & \textbf{96.4\%} & 79.4\% & \textbf{3.6\%} & 32.5\% \\
\rowcolor{lightgrayrow}
InsCAT 
& \textbf{98.2\%} & 95.0\% & \textbf{96.6\%} & 5.0\% & \textbf{1.8\%} \\
\bottomrule
\end{tabular}
}
  \vspace{-4 mm} 
\end{table}

Tab.~\ref{tab:physical} summarizes the physical-world detection results. The undefended model maintains a high precision of 97.8\% and a low FPR of 2.2\%, but its recall is limited to 78.2\%, resulting in an ASR of 21.8\%. AT-Mix substantially improves recall to 96.4\% and reduces ASR to 3.6\%, but its precision drops to 67.5\% and its FPR increases sharply to 32.5\%, showing that the improved detection rate is accompanied by severe texture-induced false positives. InsCAT achieves the best overall balance, with the highest precision of 98.2\%, an F1-score of 96.6\%, and the lowest FPR of 1.8\%, while maintaining a high recall of 95.0\% and a low ASR of 5.0\%. Compared with the undefended model, InsCAT reduces ASR by 16.8 percentage points and further lowers FPR from 2.2\% to 1.8\%, demonstrating improved physical-world robustness without introducing texture-triggered false detections.

Fig.~\ref{fig:physical} further provides qualitative detection and Grad-CAM results under different viewpoints. For both AdvReal and AdvSerial, the undefended model produces low-confidence or missed detections and concentrates its responses on local adversarial texture regions, whereas InsCAT maintains stable person detections across viewpoints and attends to broader body regions. The handheld-texture results further reveal the shortcut behavior of AT-Mix, which incorrectly detects isolated texture clothes as persons. In contrast, InsCAT suppresses these texture-triggered responses while preserving the detection of the actual person, consistent with its substantially lower physical-world FPR.

\begin{figure}[t]
  \centering
  \includegraphics[width=8.6cm]{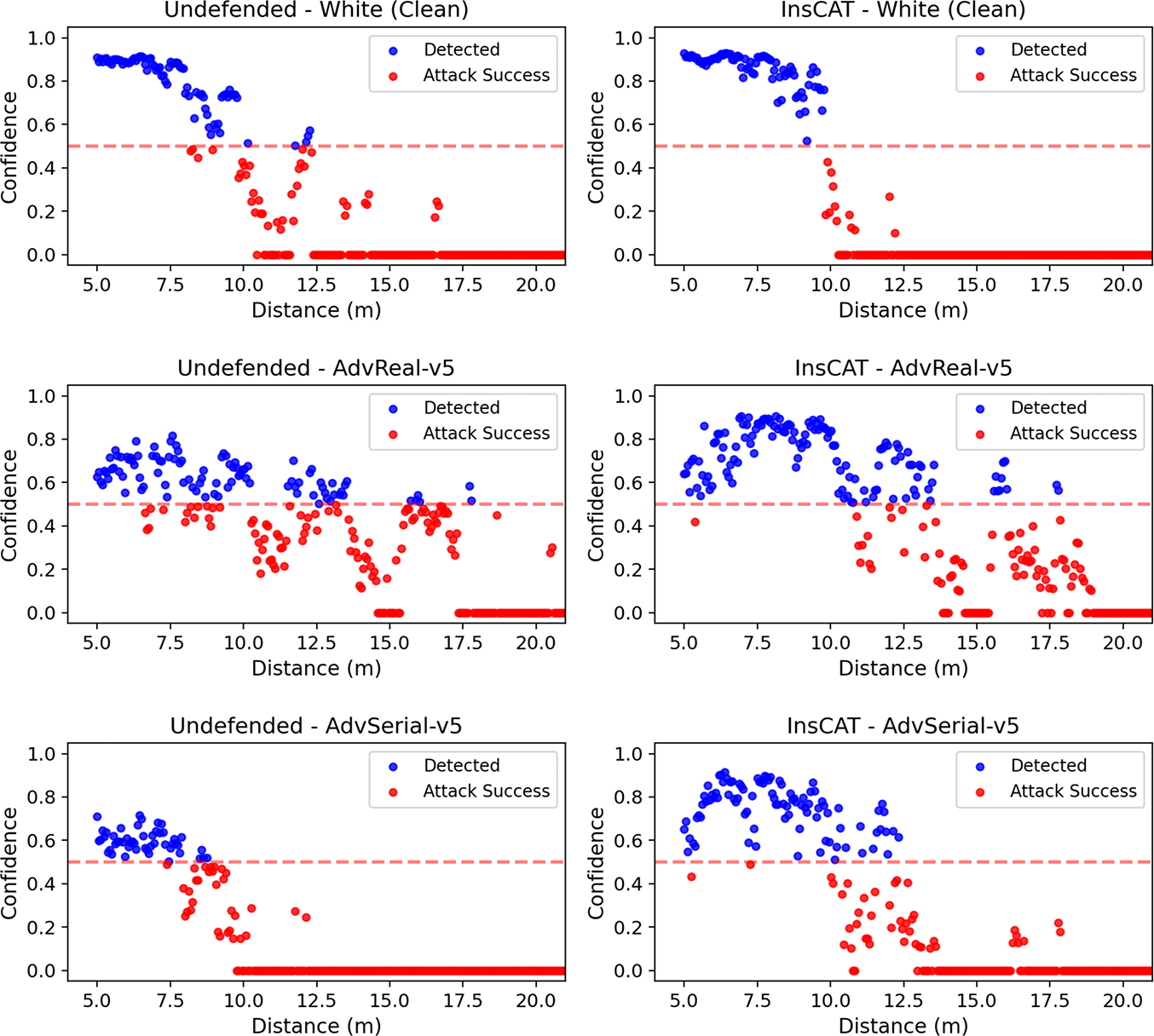}
  \caption{Detection confidence versus distance for the undefended model and InsCAT under clean, AdvReal-v5, and AdvSerial-v5 conditions. The dashed line denotes the confidence threshold of 0.5.}
  \label{fig:distance}
  \vspace{-4 mm} 
\end{figure}

\begin{figure*}[h!]
  \centering
  \includegraphics[width=15cm]{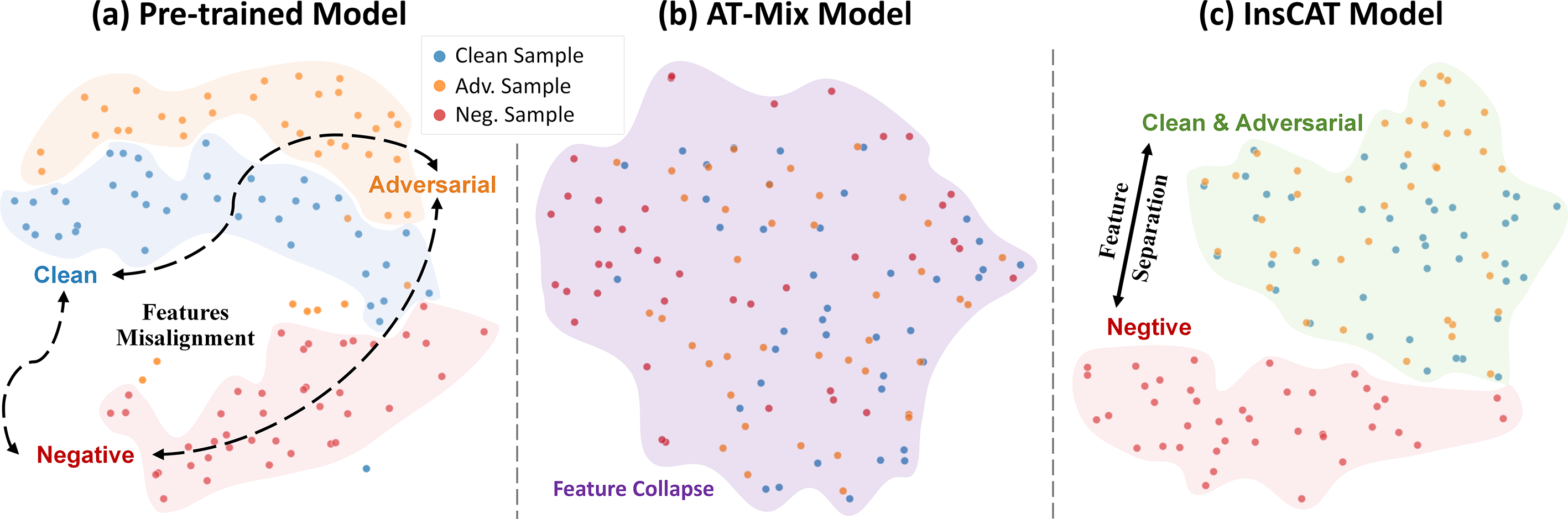}
  \caption{t-SNE visualisation of clean person, adversarial person, and texture-only RoI features for the pretrained model, AT-Mix, and InsCAT.}
  \label{fig:tsne}
  \vspace{-4 mm} 
\end{figure*}

\textbf{Performance at different distances.} To further analyse the effect of viewing distance, we recorded five videos for each garment, including the clean white garment, AdvReal-v5, and AdvSerial-v5, in which the subject continuously walked from the far field towards the camera. We computed the frame-level person confidence and averaged the results over the five videos for each garment, as shown in Fig.~\ref{fig:distance}. In the practically relevant near-distance range of 5-10 m, InsCAT consistently maintains high detection confidence across all three garments, reaching average confidence values of 0.76-0.83, whereas the undefended detector drops to 0.49-0.62 under adversarial textures. AdvSerial-v5 causes the strongest degradation, reducing the undefended model to only 12.5\% recall over the full sequences; nevertheless, InsCAT increases the recall to 23.8\% and, in the 10-15 m range, raises the average confidence from 0.012 to 0.231. Beyond approximately 15 m, both models exhibit substantial confidence collapse. A similar degradation is also observed for the clean white garment, indicating that the far-distance failures are mainly caused by the reduced apparent target scale rather than adversarial textures alone. These results demonstrate that InsCAT improves physical-world robustness primarily in the near-to-middle distance range where the pedestrian remains visually resolvable, while preserving more reliable person detection under strong adversarial textures.

\subsection{Explainability}
\label{sec:explainability}

To examine how different training strategies affect feature representations, we extract mid-level RoI features from clean person samples, adversarial person samples, and human-free texture-only negative samples, and visualize their distributions using t-SNE. The same samples and feature extraction settings are used for the pre-trained model, AT-Mix, and InsCAT to ensure a consistent comparison.

The physical Grad-CAM results in Fig.~\ref{fig:physical} provide image-space evidence of where each detector focuses in real scenes. AT-Mix exhibits strong responses to the adversarial texture itself, including isolated texture clothes without a person, whereas InsCAT shifts its attention toward broader human-body regions and suppresses texture-triggered detections. The following t-SNE analysis complements this observation by examining how clean persons, adversarial persons, and texture-only negatives are organized in the learned feature space.

As shown in Fig.~\ref{fig:tsne}, the pre-trained model exhibits a clear distribution shift between clean and adversarial person features, indicating that adversarial textures substantially alter the original person representation. After AT-Mix training, clean, adversarial, and texture-only features become strongly mixed, suggesting that the detector incorporates adversarial texture cues into the person representation. In contrast, InsCAT brings clean and adversarial person features into a shared feature region while maintaining clear separation from texture-only negatives. This distribution is consistent with the objective of SICA and the reduced texture FPR observed in Fig.~\ref{fig:texture_shortcut}. Together with the physical attention patterns in Fig.~\ref{fig:physical}, these results indicate that InsCAT improves the consistency of person representations under adversarial perturbations while reducing the detector's dependence on texture-only cues.

\section{Discussion and Limitations}
\label{sec:discussion}

\subsection{What Does the Detector Learn Wrong?}

The detector can learn that adversarial texture itself is evidence for the person category. This association arises because the texture repeatedly appears on positive person instances during adversarial training. The resulting model may detect a textured person successfully while continuing to predict a person after the human body has been removed. The learned error therefore lies not in the final prediction itself, but in the evidence used to support the prediction. The detector associates person presence with the attack texture instead of requiring sufficient person related information.

AT-Mix provides a clear example. It achieves the highest adversarial AP$_{50}$ of $0.802$, while its texture similarity remains $0.333$ and its similarity gap is only $0.124$. On human free texture inputs, it produces an average texture FPR of $46.9\%$. InsCAT retains an adversarial AP$_{50}$ of $0.773$, while reducing texture similarity to $0.140$, increasing the similarity gap to $0.239$, and lowering texture FPR to $7.3\%$. The physical experiments show the same pattern. The AT-Mix model reaches $96.4\%$ recall with $67.5\%$ precision and a $32.5\%$ FPR. InsCAT maintains $95.0\%$ recall while increasing precision to $98.2\%$ and reducing FPR to $1.8\%$.

The results reveal an important distinction for physical adversarial robustness. Adversarial AP measures whether the target remains detectable when the person and attack texture appear together. Texture FPR and feature similarity indicate whether the texture has become independent evidence for the target category. A complete evaluation should therefore examine detection performance together with whether the learned visual cue remains consistent with the intended object evidence.

\subsection{How Should Physical Adversarial Training Correct It?}

The ablation shows that shortcut suppression depends on the joint design of attack generation, feature learning, and optimisation regulation. Pull increases the similarity gap from $0.119$ to $0.200$ and reduces texture similarity from $0.317$ to $0.303$. This establishes the clean person instance as the positive reference. Adding Push without Guard contracts the gap to $0.016$ and raises texture similarity to $0.328$. The feature constraint becomes poorly coordinated when the training trajectory does not maintain a suitable balance among clean detection, adversarial detection, and contrastive alignment.

Full InsCAT restores this coordination. Compared with Pull and Push alone, it increases the similarity gap from $0.016$ to $0.239$, reduces texture similarity from $0.328$ to $0.140$, and raises adversarial AP$_{50}$ from $0.739$ to $0.773$. SICA defines the desired relationship among clean person, adversarial person, and texture only features. Guard maintains the optimisation emphasis required for this relationship to emerge. ROPO continually updates the shared texture against the current detector and supplies matched samples from an evolving attack trajectory.

The results indicate that robust physical adversarial training requires controlling not only the strength of the adversarial signal, but also the evidence that the detector associates with the target category. ROPO maintains an evolving adversarial signal, SICA controls the relationship among clean person, adversarial person, and texture only features, and Guard stabilises this relationship throughout training. The consistent improvements on YOLOv5n, Faster R-CNN, and DN-DETR show that this design remains effective across one-stage, two-stage, and Transformer-based detectors. Cross attack robustness therefore depends on both the attacks presented during training and the evidence that the detector is encouraged to associate with the target category.

\subsection{Broader Implications and Limitations}

The findings suggest that reliability-oriented defence design for AI-based perception systems should account for how adversarial signals are embedded in target instances. InsCAT reaches an average attack AP of $82.3\%$ on the rendered nuScenes domain and exceeds the strongest competing method by $11.1$ points. Its advantage is most evident when the adversarial texture interacts with clothing geometry, body pose, and viewpoint composition. In such settings, the texture becomes part of the target appearance, making feature level regulation more suitable than treating the attack as an isolated image region.

The results also highlight the importance of evaluating both robustness and the evidence supporting it. High adversarial AP can coexist with strong texture dependence, while low texture FPR and reduced feature similarity reveal whether the attack texture remains an independent cue for the target category. The distinction provides a useful evaluation principle for future physical adversarial defences. InsCAT further preserves direct detector inference with a latency of $0.29$ ms per image, showing that shortcut resistant training can improve the learned representation without adding a deployment time processing stage.

The present study focuses on RGB person detection and wearable textures generated from a finite collection of garments, motions, and rendering conditions. Broader scene variation in intelligent transportation environments, additional target categories, scale-aware detection, and adaptive attacks against the learned representation provide natural extensions. The directions can further examine how shortcut resistant feature learning transfers across physical environments and perception tasks.

\section{Conclusion}
\label{sec:conclusion}

This study shows that the central weakness of adversarial training against physically realizable person attacks lies in what the detector learns from patched positive samples. Repeated co-occurrence makes adversarial texture an independent cue for person presence, allowing high detection performance under attack to coexist with texture-triggered false detections and limited transfer to new attacks. InsCAT combines ROPO, SICA, and Guard to maintain an evolving adversary, regulate the relationship among clean person, adversarial person, and texture-only features, and coordinate the training trajectory. On the rendered nuScenes domain, InsCAT achieves an average attack AP of $82.3\%$, exceeding the strongest competing method by $11.1$ points, while reducing texture FPR from $46.9\%$ to $7.3\%$ relative to AT-Mix. Consistent improvements across three detector families demonstrate the general applicability of the proposed feature regulation. Physical evaluations further verify the reliability of InsCAT under realistic conditions, while its direct inference design avoids additional deployment overhead. These findings establish that physical adversarial robustness depends on both maintaining target detection under attack and preventing adversarial texture from becoming independent evidence for the target category. The proposed framework provides a reliability enhancement approach for AI-based perception systems deployed in intelligent transportation infrastructure and other safety-critical environments.

\bibliographystyle{elsarticle-num-names} 
\bibliography{cas-refs}

@INPROCEEDINGS{huang2024advswap,
  author={Huang, Yuanhao and Zhang, Qinfan and Xing, Jiandong and Cheng, Mengyue and Yu, Haiyang and Ren, Yilong and Xiong, Xiao},
  booktitle={2024 IEEE 27th International Conference on Intelligent Transportation Systems (ITSC)}, 
  title={Advswap: Covert Adversarial Perturbation with High Frequency Info-Swapping for Autonomous Driving Perception}, 
  year={2024},
  volume={},
  number={},
  pages={1686-1693},
  doi={10.1109/ITSC58415.2024.10920187}}

@inproceedings{xu2020adversarial,
  title={Adversarial t-shirt! evading person detectors in a physical world},
  author={Xu, Kaidi and Zhang, Gaoyuan and Liu, Sijia and Fan, Quanfu and Sun, Mengshu and Chen, Hongge and Chen, Pin-Yu and Wang, Yanzhi and Lin, Xue},
  booktitle={European conference on computer vision},
  pages={665--681},
  year={2020},
  organization={Springer}
}

@article{tom2017adversarial,
  author       = {Tom B. Brown and
                  Dandelion Man{\'{e}} and
                  Aurko Roy and
                  Mart{\'{\i}}n Abadi and
                  Justin Gilmer},
  title        = {Adversarial Patch},
  journal      = {CoRR},
  volume       = {abs/1712.09665},
  year         = {2017},
  url          = {http://arxiv.org/abs/1712.09665},
  eprinttype   = {arXiv},
  eprint       = {1712.09665},
  bibsource    = {dblp computer science bibliography, https://dblp.org}
}

@inproceedings{hu2022adversarial,
  title={Adversarial texture for fooling person detectors in the physical world},
  author={Hu, Zhanhao and Huang, Siyuan and Zhu, Xiaopei and Sun, Fuchun and Zhang, Bo and Hu, Xiaolin},
  booktitle={Proceedings of the IEEE/CVF conference on computer vision and pattern recognition},
  pages={13307--13316},
  year={2022}
}

@inproceedings{hu2023physically,
  title={Physically realizable natural-looking clothing textures evade person detectors via 3d modeling},
  author={Hu, Zhanhao and Chu, Wenda and Zhu, Xiaopei and Zhang, Hui and Zhang, Bo and Hu, Xiaolin},
  booktitle={Proceedings of the IEEE/CVF conference on computer vision and pattern recognition},
  pages={16975--16984},
  year={2023}
}

@inproceedings{kumar2025unified,
  title={A unified, resilient, and explainable adversarial patch detector},
  author={Kumar, Vishesh and Agarwal, Akshay},
  booktitle={Proceedings of the Computer Vision and Pattern Recognition Conference},
  pages={30387--30397},
  year={2025}
}

@article{madry2017towards,
  title={Towards deep learning models resistant to adversarial attacks},
  author={Madry, Aleksander and Makelov, Aleksandar and Schmidt, Ludwig and Tsipras, Dimitris and Vladu, Adrian},
  journal={arXiv preprint arXiv:1706.06083},
  year={2017}
}

@inproceedings{zhang2019towards,
  title={Towards adversarially robust object detection},
  author={Zhang, Haichao and Wang, Jianyu},
  booktitle={Proceedings of the IEEE/CVF International Conference on Computer Vision},
  pages={421--430},
  year={2019}
}

@inproceedings{li2025pbcat,
  title={Pbcat: Patch-based composite adversarial training against physically realizable attacks on object detection},
  author={Li, Xiao and Zhu, Yiming and Huang, Yifan and Zhang, Wei and He, Yingzhe and Shi, Jie and Hu, Xiaolin},
  booktitle={Proceedings of the IEEE/CVF International Conference on Computer Vision},
  pages={24456--24466},
  year={2025}
}

@article{geirhos2020shortcut,
  title={Shortcut learning in deep neural networks},
  author={Geirhos, Robert and Jacobsen, J{\"o}rn-Henrik and Michaelis, Claudio and Zemel, Richard and Brendel, Wieland and Bethge, Matthias and Wichmann, Felix A},
  journal={Nature Machine Intelligence},
  volume={2},
  number={11},
  pages={665--673},
  year={2020},
  publisher={Nature Publishing Group UK London}
}

@article{shah2020pitfalls,
  title={The pitfalls of simplicity bias in neural networks},
  author={Shah, Harshay and Tamuly, Kaustav and Raghunathan, Aditi and Jain, Prateek and Netrapalli, Praneeth},
  journal={Advances in Neural Information Processing Systems},
  volume={33},
  pages={9573--9585},
  year={2020}
}

@article{bayer2026higher,
  title={Higher-Order Adversarial Patches for Real-Time Object Detectors},
  author={Bayer, Jens and Becker, Stefan and M{\"u}nch, David and Arens, Michael and Beyerer, J{\"u}rgen},
  journal={arXiv preprint arXiv:2601.04991},
  year={2026}
}

@inproceedings{strack2024defending,
  title={Defending against physical adversarial patch attacks on infrared human detection},
  author={Strack, Lukas and Waseda, Futa and Nguyen, Huy H and Zheng, Yinqiang and Echizen, Isao},
  booktitle={2024 IEEE International Conference on Image Processing (ICIP)},
  pages={3896--3902},
  year={2024},
  organization={IEEE}
}

@article{ilyas2019adversarial,
  title={Adversarial examples are not bugs, they are features},
  author={Ilyas, Andrew and Santurkar, Shibani and Tsipras, Dimitris and Engstrom, Logan and Tran, Brandon and Madry, Aleksander},
  journal={Advances in neural information processing systems},
  volume={32},
  year={2019}
}

@article{kurakin2016adversarial,
  title={Adversarial machine learning at scale},
  author={Kurakin, Alexey and Goodfellow, Ian and Bengio, Samy},
  journal={arXiv preprint arXiv:1611.01236},
  year={2016}
}

@article{huang2026advreal,
  title={AdvReal: Physical adversarial patch generation framework for security evaluation of object detection systems},
  author={Huang, Yuanhao and Ren, Yilong and Wang, Jinlei and Huo, Lujia and Bai, Xuesong and Zhang, Jinchuan and Yu, Haiyang},
  journal={Expert Systems with Applications},
  volume={296},
  pages={128967},
  year={2026},
  publisher={Elsevier}
}

@inproceedings{thys2019fooling,
  title={Fooling automated surveillance cameras: adversarial patches to attack person detection},
  author={Thys, Simen and Van Ranst, Wiebe and Goedem{\'e}, Toon},
  booktitle={Proceedings of the IEEE/CVF conference on computer vision and pattern recognition workshops},
  pages={0--0},
  year={2019}
}

@article{wang2025transferable,
  title={Transferable and robust dynamic adversarial attack against object detection models},
  author={Wang, Xinxin and Chen, Jing and Zhang, Zijun and He, Kun and Wu, Zongru and Du, Ruiying and Li, Qiao and Liu, Gongshen},
  journal={IEEE Internet of Things Journal},
  volume={12},
  number={11},
  pages={16171--16180},
  year={2025},
  publisher={IEEE}
}

@inproceedings{hu2021naturalistic,
  title={Naturalistic physical adversarial patch for object detectors},
  author={Hu, Yu-Chih-Tuan and Kung, Bo-Han and Tan, Daniel Stanley and Chen, Jun-Cheng and Hua, Kai-Lung and Cheng, Wen-Huang},
  booktitle={Proceedings of the IEEE/CVF international conference on computer vision},
  pages={7848--7857},
  year={2021}
}

@inproceedings{huang2023t,
  title={T-sea: Transfer-based self-ensemble attack on object detection},
  author={Huang, Hao and Chen, Ziyan and Chen, Huanran and Wang, Yongtao and Zhang, Kevin},
  booktitle={Proceedings of the IEEE/CVF conference on computer vision and pattern recognition},
  pages={20514--20523},
  year={2023}
}

@inproceedings{naseer2019local,
  title={Local gradients smoothing: Defense against localized adversarial attacks},
  author={Naseer, Muzammal and Khan, Salman and Porikli, Fatih},
  booktitle={2019 IEEE winter conference on applications of computer vision (WACV)},
  pages={1300--1307},
  year={2019},
  organization={IEEE}
}

@inproceedings{tarchoun2023jedi,
  title={Jedi: Entropy-based localization and removal of adversarial patches},
  author={Tarchoun, Bilel and Ben Khalifa, Anouar and Mahjoub, Mohamed Ali and Abu-Ghazaleh, Nael and Alouani, Ihsen},
  booktitle={Proceedings of the IEEE/CVF conference on computer vision and pattern recognition},
  pages={4087--4095},
  year={2023}
}

@inproceedings{jing2024pad,
  title={PAD: Patch-agnostic defense against adversarial patch attacks},
  author={Jing, Lihua and Wang, Rui and Ren, Wenqi and Dong, Xin and Zou, Cong},
  booktitle={Proceedings of the IEEE/CVF Conference on Computer Vision and Pattern Recognition},
  pages={24472--24481},
  year={2024}
}

@inproceedings{liu2022segment,
  title={Segment and complete: Defending object detectors against adversarial patch attacks with robust patch detection},
  author={Liu, Jiang and Levine, Alexander and Lau, Chun Pong and Chellappa, Rama and Feizi, Soheil},
  booktitle={Proceedings of the IEEE/CVF conference on computer vision and pattern recognition},
  pages={14973--14982},
  year={2022}
}

@inproceedings{xu2023patchzero,
  title={Patchzero: Defending against adversarial patch attacks by detecting and zeroing the patch},
  author={Xu, Ke and Xiao, Yao and Zheng, Zhaoheng and Cai, Kaijie and Nevatia, Ram},
  booktitle={Proceedings of the IEEE/CVF Winter Conference on Applications of Computer Vision},
  pages={4632--4641},
  year={2023}
}

@inproceedings{wu2024napguard,
  title={Napguard: Towards detecting naturalistic adversarial patches},
  author={Wu, Siyang and Wang, Jiakai and Zhao, Jiejie and Wang, Yazhe and Liu, Xianglong},
  booktitle={Proceedings of the IEEE/CVF Conference on Computer Vision and Pattern Recognition},
  pages={24367--24376},
  year={2024}
}

@article{shafahi2019adversarial,
  title={Adversarial training for free!},
  author={Shafahi, Ali and Najibi, Mahyar and Ghiasi, Mohammad Amin and Xu, Zheng and Dickerson, John and Studer, Christoph and Davis, Larry S and Taylor, Gavin and Goldstein, Tom},
  journal={Advances in neural information processing systems},
  volume={32},
  year={2019}
}

@article{wu2019defending,
  title={Defending against physically realizable attacks on image classification},
  author={Wu, Tong and Tong, Liang and Vorobeychik, Yevgeniy},
  journal={arXiv preprint arXiv:1909.09552},
  year={2019}
}

@inproceedings{rao2020adversarial,
  title={Adversarial training against location-optimized adversarial patches},
  author={Rao, Sukrut and Stutz, David and Schiele, Bernt},
  booktitle={European conference on computer vision},
  pages={429--448},
  year={2020},
  organization={Springer}
}

@article{metzen2021meta,
  title={Meta adversarial training against universal patches},
  author={Metzen, Jan Hendrik and Finnie, Nicole and Hutmacher, Robin},
  journal={arXiv preprint arXiv:2101.11453},
  year={2021}
}

@article{xiao2020noise,
  title={Noise or signal: The role of image backgrounds in object recognition},
  author={Xiao, Kai and Engstrom, Logan and Ilyas, Andrew and Madry, Aleksander},
  journal={arXiv preprint arXiv:2006.09994},
  year={2020}
}

@inproceedings{geirhos2018imagenet,
  title={ImageNet-trained CNNs are biased towards texture; increasing shape bias improves accuracy and robustness},
  author={Geirhos, Robert and Rubisch, Patricia and Michaelis, Claudio and Bethge, Matthias and Wichmann, Felix A and Brendel, Wieland},
  booktitle={International conference on learning representations},
  year={2018}
}

@article{burgert2026imagenet,
  title={ImageNet-trained CNNs are not biased towards texture: Revisiting feature reliance through controlled suppression},
  author={Burgert, Tom and Stoll, Oliver and Rota, Paolo and Demir, Begum},
  journal={Advances in Neural Information Processing Systems},
  volume={38},
  pages={60809--60830},
  year={2026}
}

@inproceedings{chen2020simple,
  title={A simple framework for contrastive learning of visual representations},
  author={Chen, Ting and Kornblith, Simon and Norouzi, Mohammad and Hinton, Geoffrey},
  booktitle={International conference on machine learning},
  pages={1597--1607},
  year={2020},
  organization={PmLR}
}

@inproceedings{he2020momentum,
  title={Momentum contrast for unsupervised visual representation learning},
  author={He, Kaiming and Fan, Haoqi and Wu, Yuxin and Xie, Saining and Girshick, Ross},
  booktitle={Proceedings of the IEEE/CVF conference on computer vision and pattern recognition},
  pages={9729--9738},
  year={2020}
}

@article{jiang2020robust,
  title={Robust pre-training by adversarial contrastive learning},
  author={Jiang, Ziyu and Chen, Tianlong and Chen, Ting and Wang, Zhangyang},
  journal={Advances in neural information processing systems},
  volume={33},
  pages={16199--16210},
  year={2020}
}

@article{kim2020adversarial,
  title={Adversarial self-supervised contrastive learning},
  author={Kim, Minseon and Tack, Jihoon and Hwang, Sung Ju},
  journal={Advances in neural information processing systems},
  volume={33},
  pages={2983--2994},
  year={2020}
}

@article{fan2021does,
  title={When does contrastive learning preserve adversarial robustness from pretraining to finetuning?},
  author={Fan, Lijie and Liu, Sijia and Chen, Pin-Yu and Zhang, Gaoyuan and Gan, Chuang},
  journal={Advances in neural information processing systems},
  volume={34},
  pages={21480--21492},
  year={2021}
}

@inproceedings{sun2021fsce,
  title={Fsce: Few-shot object detection via contrastive proposal encoding},
  author={Sun, Bo and Li, Banghuai and Cai, Shengcai and Yuan, Ye and Zhang, Chi},
  booktitle={Proceedings of the IEEE/CVF conference on computer vision and pattern recognition},
  pages={7352--7362},
  year={2021}
}

@inproceedings{caesar2020nuscenes,
  title={nuscenes: A multimodal dataset for autonomous driving},
  author={Caesar, Holger and Bankiti, Varun and Lang, Alex H and Vora, Sourabh and Liong, Venice Erin and Xu, Qiang and Krishnan, Anush and Pan, Yu and Baldan, Giancarlo and Beijbom, Oscar},
  booktitle={Proceedings of the IEEE/CVF conference on computer vision and pattern recognition},
  pages={11621--11631},
  year={2020}
}

@article{metzen2017detecting,
  title={On detecting adversarial perturbations},
  author={Metzen, Jan Hendrik and Genewein, Tim and Fischer, Volker and Bischoff, Bastian},
  journal={arXiv preprint arXiv:1702.04267},
  year={2017}
}

@inproceedings{xu2021using,
  title={Using feature alignment can improve clean average precision and adversarial robustness in object detection},
  author={Xu, Weipeng and Huang, Hongcheng and Pan, Shaoyou},
  booktitle={2021 IEEE International Conference on Image Processing (ICIP)},
  pages={2184--2188},
  year={2021},
  organization={IEEE}
}

@inproceedings{lin2014microsoft,
  title={Microsoft {COCO}: Common objects in context},
  author={Lin, Tsung-Yi and Maire, Michael and Belongie, Serge and Hays, James and Perona, Pietro and Ramanan, Deva and Doll{\'a}r, Piotr and Zitnick, C Lawrence},
  booktitle={Computer Vision--ECCV 2014: 13th European Conference, Zurich, Switzerland, September 6-12, 2014, Proceedings, Part V 13},
  pages={740--755},
  year={2014},
  organization={Springer}
}

@inproceedings{inriaperson,
author = {Dalal, Navneet and Triggs, Bill},
title = {Histograms of Oriented Gradients for Human Detection},
year = {2005},
isbn = {0769523722},
publisher = {IEEE Computer Society},
address = {USA},
url = {https://doi.org/10.1109/CVPR.2005.177},
doi = {10.1109/CVPR.2005.177},
booktitle = {Proceedings of the 2005 IEEE Computer Society Conference on Computer Vision and Pattern Recognition (CVPR'05) - Volume 1 - Volume 01},
pages = {886–893},
numpages = {8},
series = {CVPR '05}
}

@article{yolov5,
  title={ultralytics/yolov5: v3. 0},
  author={Jocher, Glenn and Stoken, Alex and Borovec, Jirka and Changyu, Liu and Hogan, Adam and Diaconu, Laurentiu and Poznanski, Jake and Yu, Lijun and Rai, Prashant and Ferriday, Russ and others},
  journal={Zenodo},
  year={2020}
}

@article{ren2016fasterrcnnrealtimeobject,
  title={Faster R-CNN: Towards real-time object detection with region proposal networks},
  author={Ren, Shaoqing and He, Kaiming and Girshick, Ross and Sun, Jian},
  journal={IEEE transactions on pattern analysis and machine intelligence},
  volume={39},
  number={6},
  pages={1137--1149},
  year={2016},
  publisher={IEEE}
}

@inproceedings{li2022dn,
  title={Dn-detr: Accelerate detr training by introducing query denoising},
  author={Li, Feng and Zhang, Hao and Liu, Shilong and Guo, Jian and Ni, Lionel M and Zhang, Lei},
  booktitle={Proceedings of the IEEE/CVF conference on computer vision and pattern recognition},
  pages={13619--13627},
  year={2022}
}

@misc{huang2026advserial,
      title={AdvSerial: Physical Adversarial Attacks on Infrastructure-mounted Pedestrian Detectors via Semantic Feature Suppression}, 
      author={Yuanhao Huang and Yilong Ren and Jinlei Wang and Xuesong Bai and Jinchuan Zhang and Haiyang Yu},
      year={2026},
      eprint={2607.17069},
      archivePrefix={arXiv},
      primaryClass={cs.CV},
      url={https://arxiv.org/abs/2607.17069}, 
}

\end{document}